\documentclass[letterpaper, 10 pt, conference]{ieeeconf}  

\IEEEoverridecommandlockouts                              

\usepackage{cite}
\usepackage{graphicx,epstopdf,epsfig}
\graphicspath{{figures/}}
\DeclareGraphicsExtensions{.eps}
\usepackage{amsmath}
\usepackage{amssymb}

\usepackage{array}
\usepackage{fixltx2e}
\usepackage{url}
\usepackage{booktabs}
\usepackage{float}
\usepackage{hyperref}
\hypersetup{colorlinks,linkcolor={red},citecolor={red},urlcolor={red}}
\usepackage[caption=false,font=scriptsize,labelfont=sf,textfont=sf]{subfig}
\usepackage[linesnumbered,ruled,vlined]{algorithm2e}

\title{\LARGE \bf
	Complementary Visual Neuronal Systems Model for Collision Sensing
}

\author{
	Qinbing Fu, \IEEEmembership{Member, IEEE} and Shigang Yue, \IEEEmembership{Senior Member, IEEE}
\thanks{
	This research has been supported by the grants from EU Horizon 2020 projects STEP2DYNA (691154) and ULTRACEPT (778062). 
	The authors are with the Machine Life and Intelligence Research Centre, Guangzhou University, Guangzhou, 510006, China, 
	and also with the Computational Intelligence Lab/Lincoln Centre for Autonomous Systems, University of Lincoln, Lincoln, LN6 7TS, United Kingdom. 
	Correspondence: \url{{qifu,syue}@lincoln.ac.uk}.
	}
}

\begin{document}

\maketitle
\thispagestyle{empty}
\pagestyle{empty}

\begin{abstract}

	Inspired by insects' visual brains, this paper presents original modelling of a complementary visual neuronal systems model for real-time and robust collision sensing. 
	Two categories of wide-field motion sensitive neurons, i.e., the lobula giant movement detectors (LGMDs) in locusts and the lobula plate tangential cells (LPTCs) in flies, have been studied, intensively. 
	The LGMDs have specific selectivity to approaching objects in depth that threaten collision; 
	whilst the LPTCs are only sensitive to translating objects in horizontal and vertical directions. 
	Though each has been modelled and applied in various visual scenes including robot scenarios, little has been done on investigating their complementary functionality and selectivity when functioning together. 
	To fill this vacancy, we introduce a hybrid model combining two LGMDs (LGMD-1 and LGMD-2) with horizontally (rightward and leftward) sensitive LPTCs (LPTC-R and LPTC-L) specialising in fast collision perception. 
	With coordination and competition between different activated neurons, the proximity feature by frontal approaching stimuli can be largely sharpened up by suppressing translating and receding motions. 
	The proposed method has been implemented in ground micro-mobile robots as embedded systems. 
	The multi-robot experiments have demonstrated the effectiveness and robustness of the proposed model for frontal collision sensing, which outperforms previous single-type neuron computation methods against translating interference.

\end{abstract}


\section{INTRODUCTION}
\label{Sec: introduction}

Fast and reliable collision perception is of paramount importance for autonomous mobile machines including ground vehicles, UAVs and robots. 
In complex and dynamic environments, it is still an open challenge for mobile machines to detect imminent collision dangers, timely and robustly, without human intervention.

Generally speaking, visual collision detection methods can vary from traditional computer vision techniques, such as object-scene segmentation, estimation or classification algorithms \cite{Vehicle-Review-2015}, to specialised sensor based strategies, like RGB-D \cite{Kinect-2013} or event-driven \cite{UAV2017(LGMD1-spiking)} cameras. 
In addition, another category of effective collision detection-and-avoidance approaches originates from biological visual systems research \cite{Fu-ALife-review,Fu-TAROS(review),Serres2017(review-optic-flow)}.

In particular, insects have tiny brains but compact visual systems serving a variety of tasks including foraging, escaping from predators, chasing mates and the like \cite{Fu-ALife-review}. 
It is believed that the insect visual systems are highly robust, flexible and energy-efficient performing in the dynamic visual world, which are excellent paradigms to learn motion perception methods. 
Modelling the underlying neural processing circuits, pathways and mechanisms will undoubtedly advance potent applications in mobile intelligent machines.

Different types pf visual neurons possess specific direction selectivity (DS) and work together to decode diverse motion patterns, including movements in depth (approaching or receding), translating, spiral motion and etc. 
Specifically, two categories of wide-field motion sensitive neurons have been identified and systematically studied, that is, the LGMDs (LGMD-1 and LGMD-2) in locusts and the LPTCs in fruit fly \textit{Drosophila}, as reviewed in \cite{Fu-ALife-review}.

\begin{figure}[t!]
	\centering
	\includegraphics[width=0.4\textwidth]{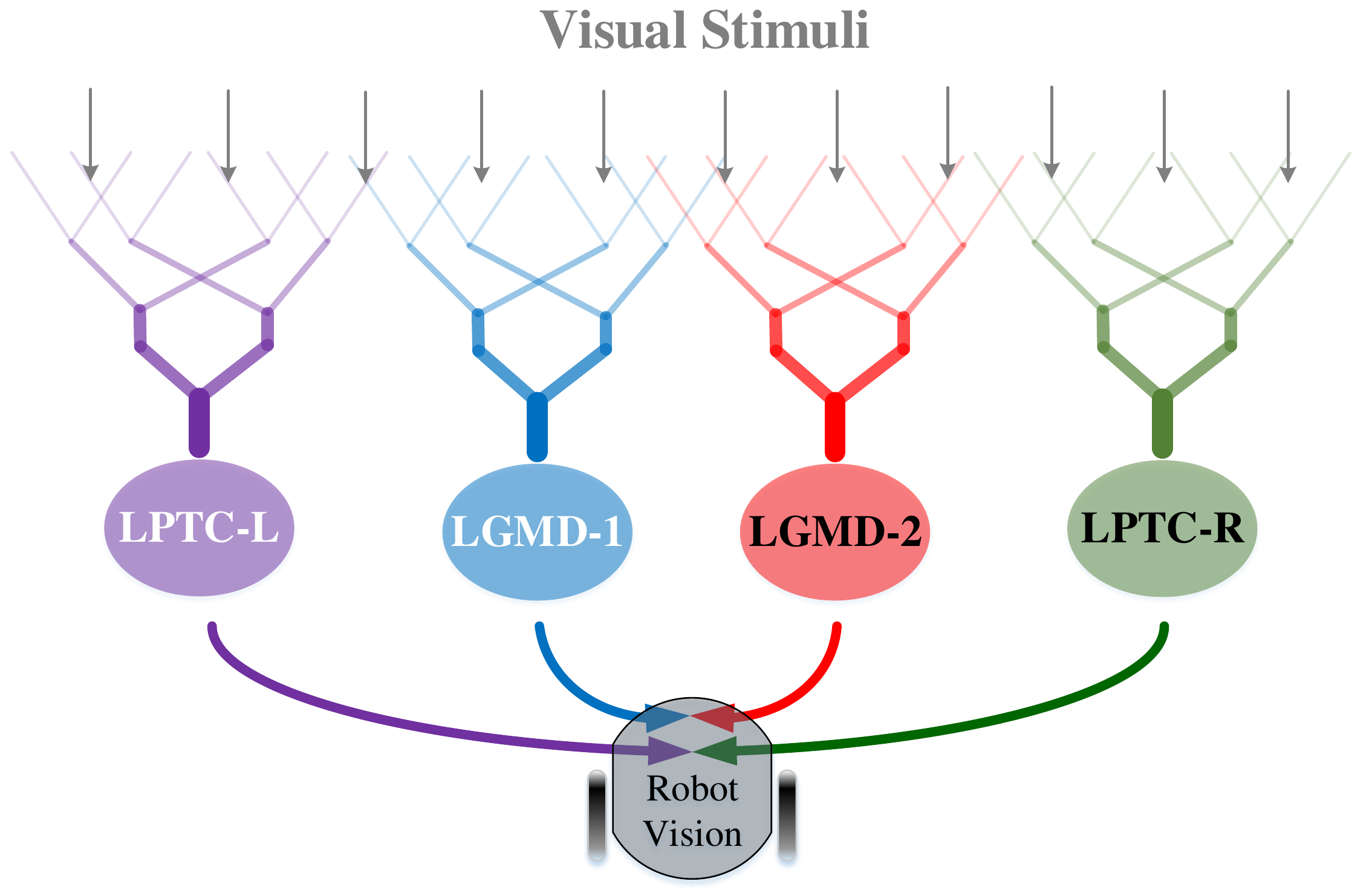}
	\caption{
		The complementary and hybrid neuronal systems model in robot embedded vision consisting of four motion sensitive neuronal circuit models functioning together for fast collision sensing with enhanced selectivity.
	}
	\label{Fig: motivation}
	\vspace{-10pt}
\end{figure}

Firstly, the two LGMDs locate in a place called `lobula area' and respond most strongly to moving objects that signal approaching rather than other sorts of movements \cite{LGMD1-1996(Rind-neural-network),Simmons-1997(LGMD2-neuron-locusts),LGMDs-2016,LGMD-2016(Gray-background-motion)}. 
Compared to the LGMD-1, the LGMD-2 possesses specific selectivity to only darker object (relative to the background) approaching. 
Either of the neurons has been investigated as quick and robust collision sensing visual systems, and successfully applied in various visual scenes including ground vehicles (e.g. \cite{LGMD-car-2017(bionic-vehicle-collision),Fu-AIAI-2019,LGMD1-ONOFF-2004,LGMD1-car-2011(risk-collision-road),LGMD1-car2007(collision-detection-cars)}), mobile robots (e.g. \cite{Badia-2010(LGMD1-nonlinear-model),Fu-2018(LGMD1-NN),Fu-2016(LGMD2-BMVC),Fu-2015(LGMD2-MLSP),Hu-SAB(hybrid),LGMD1-walking-robot,Fu-LGMD2-TCYB}) and UAVs (e.g. \cite{UAV-2018(LGMD1-Opticflow-PID),spikingLGMD-TNNLS}). 
In addition, a recent study for the first time demonstrated the coordination of both the LGMDs for robust collision sensing in different lighting conditions, from bright to dark scenarios \cite{Fu2017a(LGMDs-IROS)}. 
The superiority of combining the two LGMDs is the enhanced collision selectivity to only darker approaching stimuli, the similar situations which ground vehicles and robots are often faced with. 
However, the current LGMDs computational models are still greatly influenced by nearby or accelerating translating stimuli: the models signal collision-like response.

\begin{figure*}[t!]
	\centering
	\includegraphics[width=0.75\linewidth]{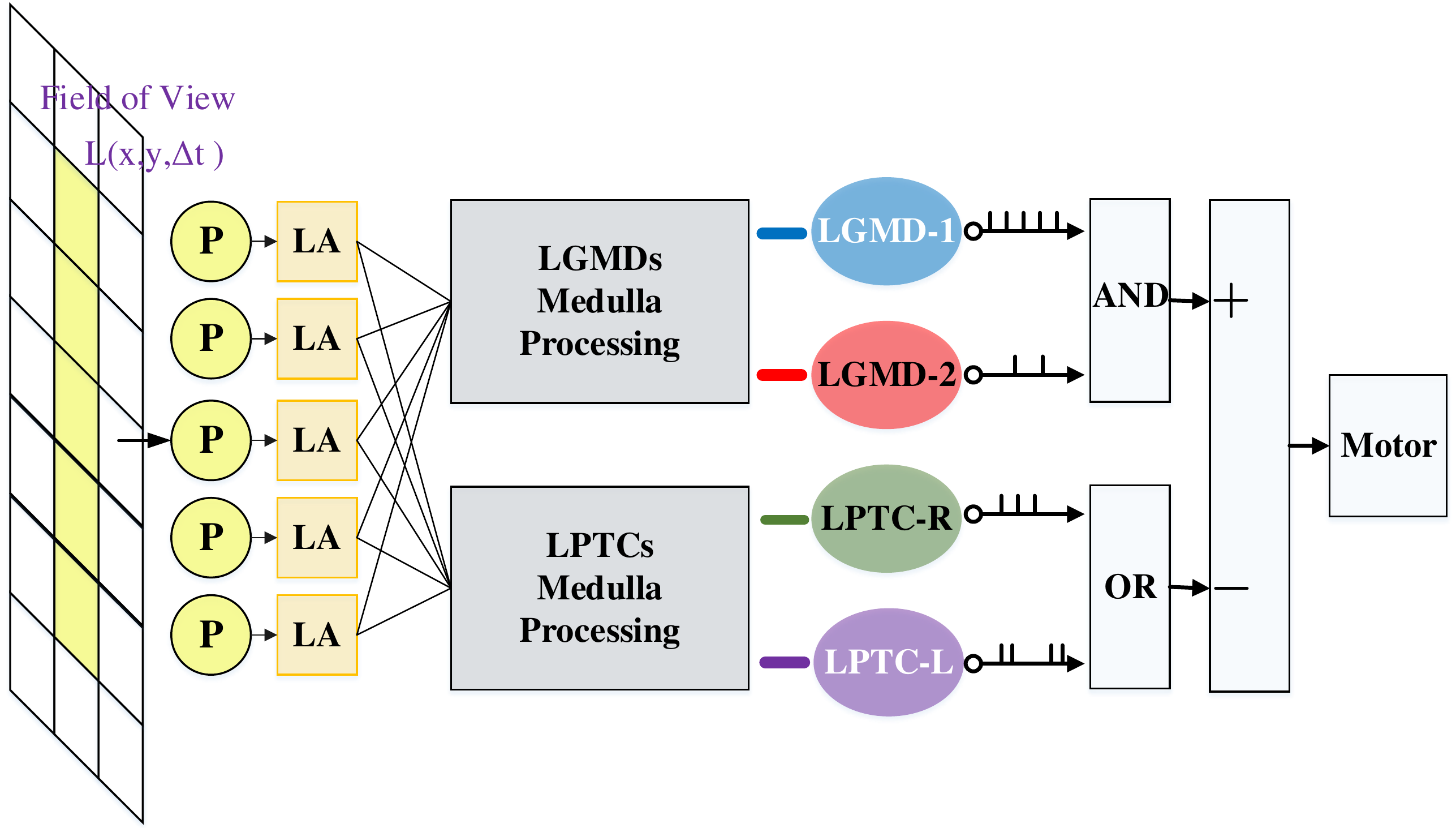}
	\caption{
		Schematic illustration of the proposed complementary hybrid neuronal systems model. 
		Taken five local optic units from the field of view to exemplify the visual processing, P and LA denote photoreceptors and lamina cells in the first two computational neuro-layers that are shared by the four spiking neurons. 
		Signal processing in the third medulla layer varies between the LGMDs and the LPTCs to generate totally different direction selectivity. 
		Finally, the spikes of four neurons with respect to time are conveyed to two logic gates, and then a comparator towards the motor.
	}
	\label{Fig: model}
	\vspace{-10pt}
\end{figure*}

On the other hand, the LPTCs have been found in the neural circuits, at the lobula plate of fruit flies \cite{Fu-ALife-review}. 
Compared to the LGMDs, a prominent feature of the LPTCs is that they are only sensitive to translating motion in four cardinal directions \cite{Maisak_2013(T4-T5-fly)}. 
For computationally implementing the LPTCs (or called fly direction selective neurons, DSNs), there are two main perspectives: 
1) the well-known bio-inspired optic flow (OF)-based approaches apply elementary motion detectors (EMDs) to reach the preliminary level of LPTCs at local optical unit level, which have been broadly used in flying robots and micro aerial vehicles, as reviewed in \cite{Nicolas-2014(Review-Fly-Robot),Serres2017(review-optic-flow)}; 
2) behind the OF-based level, a recent LPTCs model mimics the \textit{Drosophila} visual processing through multiple neuro-layers for decoding translating-object direction against cluttered backgrounds \cite{Fu2017(fly-DSNs-IJCNN)}.

Obviously, the specific functionality and DS of the LGMDs and the LPTCs can complement each other, perfectly. 
However, little has been done on investigating the potential and advantage of their coordination in motion perception. 
To fill this gap, we propose a hybrid model combining four visual neuronal system models functioning together (see Fig. \ref{Fig: motivation}), i.e., LGMD-1, LGMD-2, LPTC-R (rightward-sensitive) and LPTC-L (leftward-sensitive), concentrating on collision sensing in dynamic scenes mixed with diverse motion patterns including translating, approaching (frontal and angular) and receding stimuli. 
A main novelty is the coordination and competition between different activated neurons to alleviate the impact by strong translating movements: the activation of LPTCs will rigorously suppress both the LGMDs. 
As a result, the proposed model responds most strongly to frontal approaching targets.

The model has been implemented in embedded vision of autonomous micro-mobile robots in multi-robot arena tests. 
We have also compared its competence with two previous single-type neuron computation methods \cite{Fu-2016(LGMD2-BMVC),Fu2017a(LGMDs-IROS)}. 
Through the verification of robot experiments, we summarise the following achievements: 
1) This paper presents a novel way of coordinating multiple neuronal systems with complementary functionality to sharpen up the collision feature in dynamic environments.
2) With more motion patterns recognised by specific neurons, the proposed model outperforms the single-type neuron models for frontal collision sensing against translating interference.
 
The rest of this paper is organised as follows: 
Section \ref{Sec: method} presents the method. 
Section \ref{Sec: configuration} introduces the robot and system configuration. 
Section \ref{Sec: experiments} illustrates the experiment results. 
Section \ref{Sec: conclusion} concludes this research.


\section{METHOD}
\label{Sec: method}

Within this section, we introduce the proposed visual model, as illustrated in Fig. \ref{Fig: model}. 
In general, the model consists of four spiking neurons and corresponding three pre-synaptic neuro-layers (retina, lamina and medulla). 
Through previous experience, we have pointed out the commonalities and disparities of implementing different motion sensitive neurons \cite{Fu-ALife-review}. 
Therefore, in this modelling study, we highlight the following structures and mechanisms: 
\begin{itemize}
	\item	The four neurons can share the same spatiotemporal visual processing in the first two layers (retina and lamina), as shown in Fig. \ref{Fig: model}.
	\item	The ON-OFF channels have been verified to be essential in motion sensitive neural circuits \cite{Fu-ALife-review}. 
	The motion information is encoded in parallel ON-OFF channels where the light-on (luminance increment) and light-off (luminance decrement) responses stream into the ON and OFF channels, respectively. 
	Importantly, our previous modelling studies have also demonstrated that such a structure can not only separate the different collision selectivity between the LGMD-1 and the LGMD-2 \cite{Fu-2018(LGMD1-NN),Fu-LGMD2-TCYB}, but also implement the \textit{Drosophila} LPTCs \cite{Fu2017(fly-DSNs-IJCNN)}.
	\item	The medulla layer is the place where the complementary DS is shaped for the four different neurons, i.e., the distinct sensitivity to approaching or translating stimuli.
\end{itemize}

Noting that avoiding repetition of algorithms presentation, we just summarise formulation of the crucial complementary DS and the spiking mechanism instead of illustrating all. 
More details can be referenced in our recent studies: LGMD-1 in \cite{Fu-2018(LGMD1-NN)}, LGMD-2 in \cite{Fu-LGMD2-TCYB} and LPTCs in \cite{Fu2017(fly-DSNs-IJCNN),Fu-ROBIO-2018}.

\subsection{Computational Retina Layer}

As illustrated in Fig. \ref{Fig: model}, in the first retina layer, there are photoreceptors arranged in a 2D matrix form that capture time-varying and single-channel (grey-scale in our case) luminance. 
The preliminary motion information can be obtained by calculating the `differential image' at every two successive frames through each local unit (or image pixel).

\subsection{Computational Lamina Layer}

In the second lamina layer, we firstly apply a bio-inspired mechanism of `Difference of Gaussians' to remove redundant environmental detailed noise and achieve edge selectivity to the preliminary motion in the 2D plane. 
This also depicts an interaction between inner excitatory and outer inhibitory fields, spatially, in which the outer field is with twice size of the inner one. 
More precisely, two convolution processes go through each local lamina cell to get excitation and inhibition, and then the former is subtracted from the latter. 

After that, there are ON and OFF channels splitting the motion signals into parallel computation encoding light-on and light-off responses, respectively. 
A half-wave rectifying mechanism is applied to filter out negative inputs in the ON channels and positive inputs in the OFF channels. 
In addition to that, the negative sign is inverted in the OFF channels.

For each lamina cell, an `adaptation state' is formed by a bio-plausible fast-depolarising-slow-repolarising (FDSR) mechanism which works effectively to reduce background flickers, temporally.

\subsection{Computational Medulla Layer}

The third medulla layer is of great importance to shape the complementary DS of the four different motion sensitive visual neurons. 
As illustrated in Fig. \ref{Fig: model}, the signal processing here varies between the LGMDs and the LPTCs.

On the aspect of LGMDs, both neurons detect potential collisions by reacting to expanding edges of objects. 
As mentioned above, the LGMD-2 has the specific selectivity to darker approaching objects; whilst the LGMD-1 responds to either darker or brighter objects. 
For realising the collision selectivity, both the lateral inhibition mechanism and the ON-OFF channels are crucial \cite{Fu-ALife-review}. 
In either ON/OFF channels, there are spatiotemporal competitions between local inhibitions and excitations. 
The lateral inhibition (I) is formed by convolving surrounding and symmetrically spreading excitations (E) with temporal delay. 
That is, 
\begin{equation}
I(x,y,t) = \iiint E(u,v,s)\ W(x-u,y-v,t-s)\ d u d v d s,
\label{Eq: LGMDs-inhibition}
\end{equation}
where $W$ is a local convolution kernel. 
$x$, $y$, $t$ are spatial and temporal positions. 
After that, the excitations are cut down by the lateral inhibitions. 
That is,
\begin{equation}
S(x,y,t) = E(x,y,t) - w \cdot I(x,y,t),
\label{Eq: LGMDs-competition}
\end{equation}
where $S$ denotes the summation cells and $w$ is a local bias. 
For achieving the LGMD-2's specific selectivity to only darker objects, the weightings in the convolution kernel $W$ of the ON channels are much raised in order to gain stronger inhibitions that rigorously suppress the excitations from the ON channels.

With regard to the medulla processing of LPTCs, the specific DS to translating motion is accomplished by non-linear spatiotemporal computations according to the EMDs, at local interneuron level. 
That is, 
\begin{equation}
R(t) = X_1(t-\epsilon) \cdot X_2(t) - X_1(t) \cdot X_2(t-\epsilon),
\label{Eq: DSNs-emd}
\end{equation}
where $R$ is the output of each pairwise motion detectors in space. 
$X_1$ and $X_2$ are two adjacent motion sensitive cells, and $\epsilon$ is the temporal delay. 
As a result, the different DS between the LPTC-R and the LPTC-L can be well differentiated.

\subsection{Spiking Neurons}

After the generation of local motion signals with specific DS in the medulla layer, the four neurons integrate all from its corresponding pre-synaptic medulla layer, linearly and spatially. 
Subsequently, a sigmoid transformation is applied as the activation function, for each neuron, to generate membrane potential \cite{Fu-LGMD2-TCYB,Fu2017(fly-DSNs-IJCNN)}. 
Note that the response of both the LGMDs is normalised within $[0.5, 1)$; whilst the responses of the LPTC-R and the LPTC-L are within $[0, 1)$ and $(-1, 0]$, respectively.

As shown in Fig. \ref{Fig: model}, all four neurons spikes with respect to time: the membrane potential (U) is encoded, exponentially by an integer-valued function as the following: 
\begin{equation}
S^{spike}(t) =  \left[e^{(K_{sp} \cdot (|U(t)| - |T_{sp}|))}\right],
\label{Eq: neurons-spiking}
\end{equation}
where $K_{sp}$ and $T_{sp}$ denote a scale coefficient and a spiking threshold. 
Therefore, with digital signals as the input, more than one spikes could be produced at every discrete frame.

\subsection{Coordination and Competition}

In the proposed method, the four neurons coordinate and compete with each other to shape the frontal collision selectivity (see Fig. \ref{Fig: model}), in which the activation of LPTCs will rigorously suppress the LGMDs. 
More concretely, the two LGMDs coordinate in collision sensing with enhanced selectivity to approaching over receding stimuli. 
Only both the LGMDs are activated, a potential collision threat is recognised. 
On the other hand, if either of the LPTCs is highly activated, the translating or angular approaching feature is detected resulting in strong inhibition on both the LGMDs, immediately. 
A comparator here works effectively to convey the spikes after competition towards motor: if the LGMDs win, the avoidance behaviour will be aroused, promptly. 
Accordingly, the proximity feature by frontal approaching objects against translating motion is well sharpened up.


\section{CONFIGURATION}
\label{Sec: configuration}

In this section, we continue to introduce the system and robot configuration. 
To ease the understanding of robot implementation, Algorithm \ref{Alg: model} elucidates the whole procedures. 
More detailed model settings can refer to our related works, the LGMDs in \cite{Fu-2016(LGMD2-BMVC),Fu2017a(LGMDs-IROS)} and the LPTCs in \cite{Fu2017(fly-DSNs-IJCNN),Fu-ROBIO-2018}. 
Compared to previous methods, the LGMDs are hybrid with the horizontally sensitive LPTCs to exclude translating interference to a large extent. 
Note that the spiking threshold is set, differently, as $T_{sp} = 0.7$ for both the LGMDs, $T_{sp} = \pm 0.2$ for the LPTCs (positive/negative for LPTC-R/LPTC-L), and $K_{sp} = 4$.

\begin{algorithm}[t!]
	\label{Alg: model}
	\caption{Robot Implementation}
	\While{Power on}
	{
		Robot monocular cameral system works\;
		Model retina layer processes input images stream\;
		Model lamina layer processing\;
		Model medulla layers processing\;
		LGMDs and LPTCs integrate signals and spike\;
		Coordination and competition between neurons\;
		\eIf{
			Spike rate of LGMDs is higher
		}
		{
			Robot agent triggers avoidance behaviour\;
			Stops moving and turn around to left or right\;
			Resume going forward\;
		}
		{
			Robot agent	goes forward\;
		}
	}
	\textbf{end}\\
\end{algorithm}

\begin{figure}[t!]
	\centering
	\subfloat[arena view]{\frame{\includegraphics[width=0.2\textwidth]{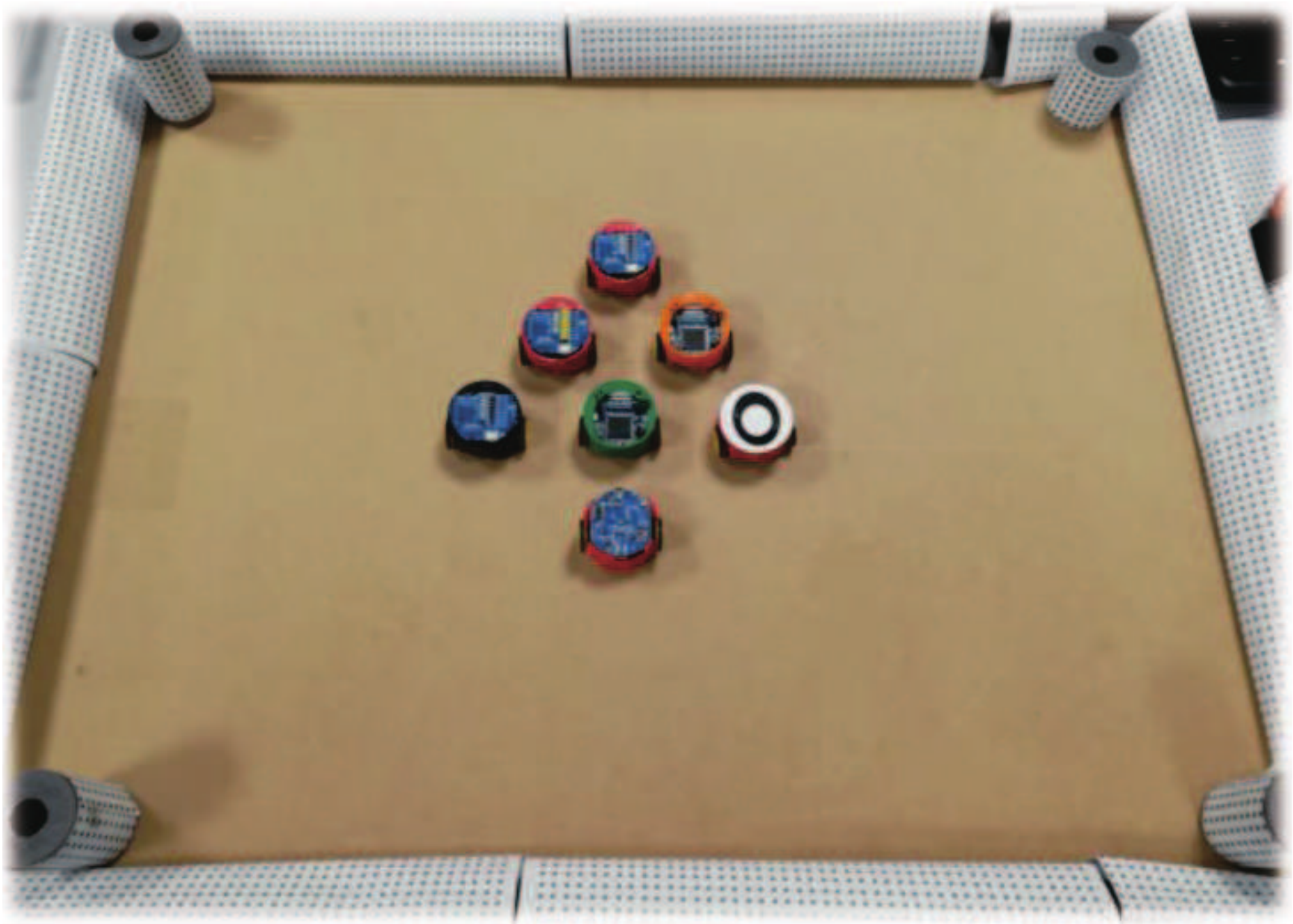}}
		\label{Fig: arena}}
	\hfil
	\subfloat[\textit{Colias} robot]{\frame{\includegraphics[width=0.265\textwidth]{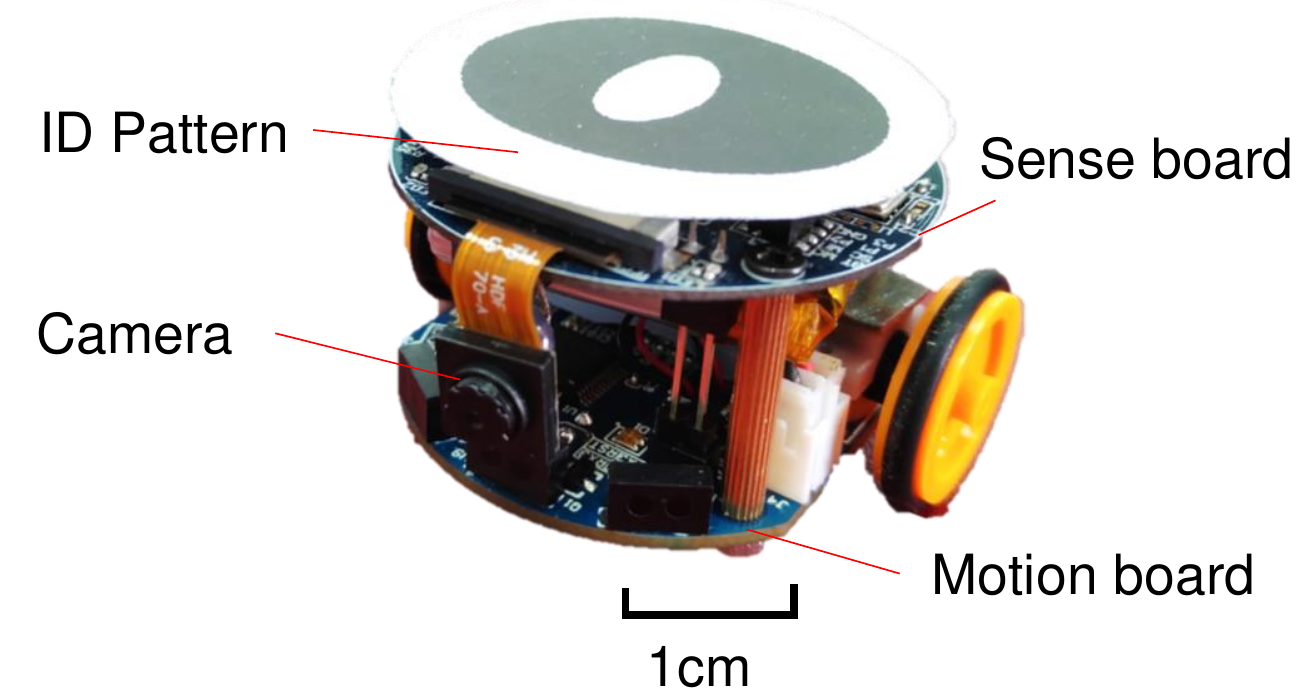}}
		\label{Fig: colias}}
	\caption{
		Illustration of the \textit{Colias} robot and the arena. 
		The ID-specific pattern on top of the robot is used to locate the robot in real time \cite{Colias-localization}.
	}
	\label{Fig: configuration}
	\vspace{-10pt}
\end{figure}

The micro-mobile robot used in this research for all the on-line experiments is called `\textit{Colias}', which is a low-cost and autonomous ground mobile platform (Fig. \ref{Fig: colias}). 
The \textit{Colias} robot has a small footprint of 4cm in diameter and 3cm in height. 
It has mainly two boards. 
The bottom motion board serves the robot with a maximum speed of roughly 35cm/s. 
Equipped with two wheels, the robot can only run on 2D surfaces. 
The upper sense board is assembled with a monocular camera (OV7670) system handling the required in-chip image processing, as the only sensor applied in this study. 
The acquired image was set at $99 \times 72$ in YUV422 format at 30 fps. 
Moreover, the visual coverage of camera could reach up to 70 degrees. 
In our implementation, the frame rate can maintain $25 \sim 40$Hz meeting the requirements for real time visual tasks. 
Furthermore, a Bluetooth device connecting the sense board was applied to get real time data, remotely, including neuronal outputs of membrane potential and spikes. 
More assembly details can be found in \cite{Hu-TAROS(Colias-IV)}.

Fig. \ref{Fig: arena} displays a top-down view of the arena used for multi-robot experiments. 
It has a small dimension of $70 \times 55 cm^2$ in acreage. 
The peripheries were decorated with textured patterns to stimulate the approaching robot.

\begin{figure*}[t!]
	\centering
	\subfloat[approaching]{\includegraphics[width=0.16\linewidth]{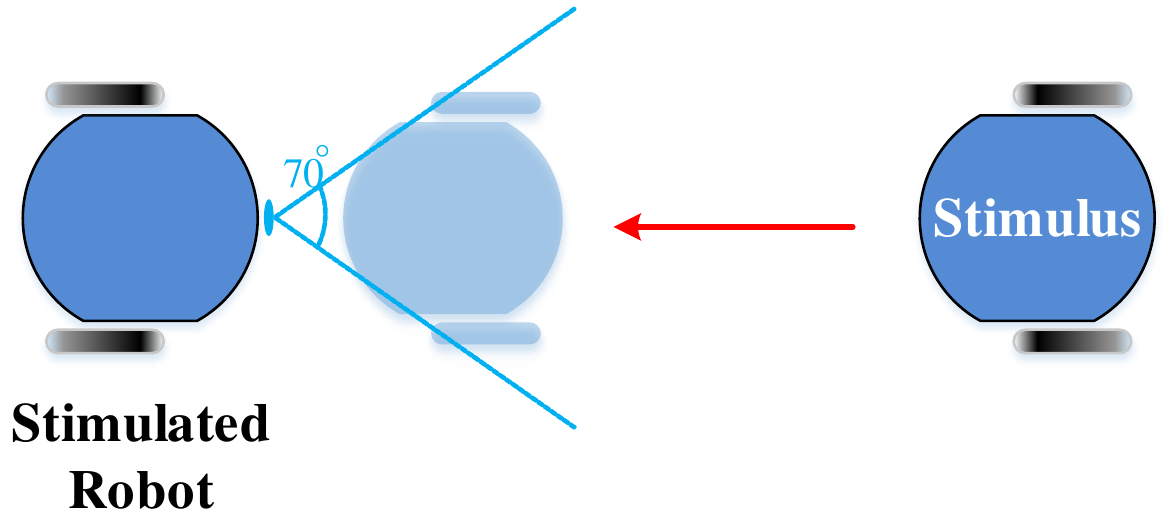}
		\label{Fig: approaching-pattern}}
	\hfil
	\hspace{0.1in}
	\subfloat[Receding]{\includegraphics[width=0.16\linewidth]{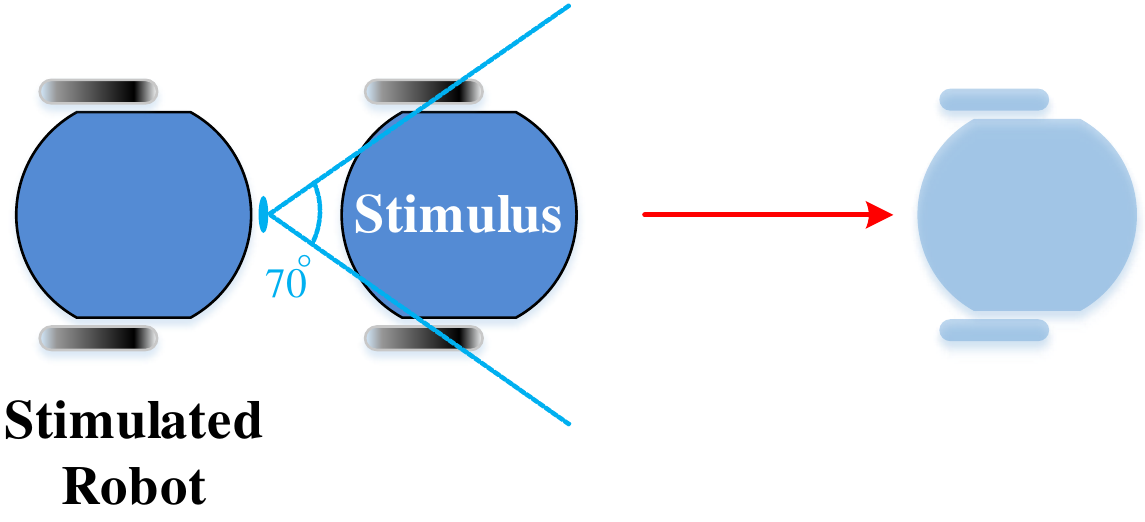}
		\label{Fig: receding-pattern}}
	\hfil
	\hspace{0.1in}
	\subfloat[Translating-R]{\includegraphics[width=0.11\linewidth]{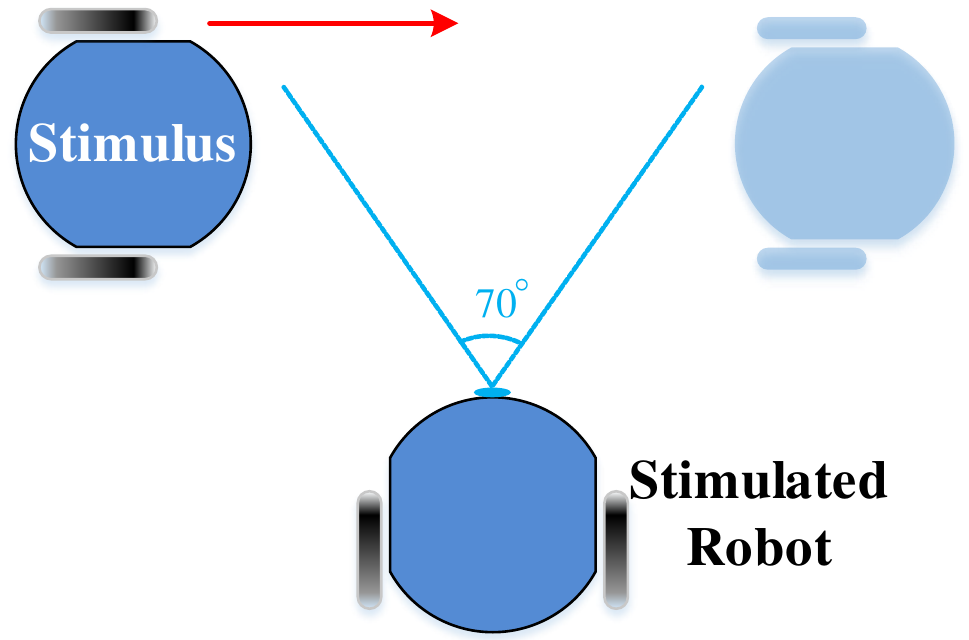}
		\label{Fig: transR-pattern}}
	\hfil
	\hspace{0.3in}
	\subfloat[Translating-L]{\includegraphics[width=0.11\linewidth]{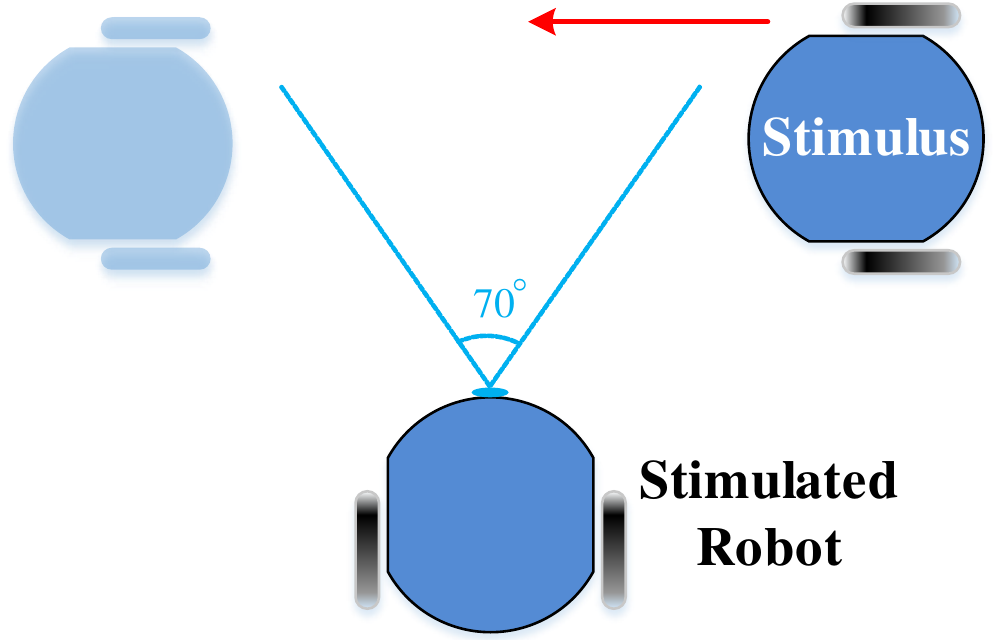}
		\label{Fig: transL-pattern}}
	\vfill
	\vspace{-0.1in}
	\subfloat{\includegraphics[width=0.24\linewidth]{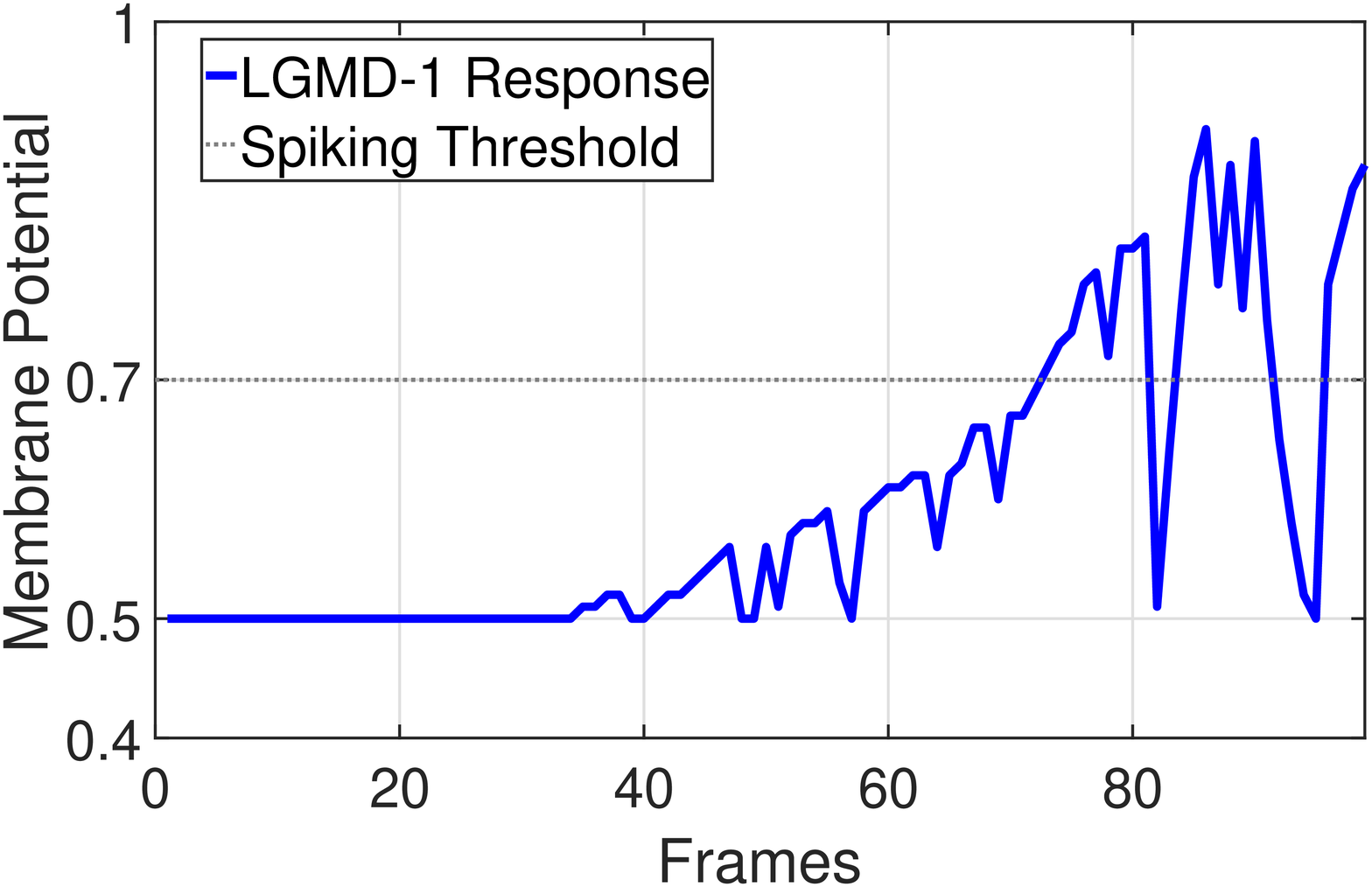}
		\label{Fig: approaching-lgmd1}}
	\hfil
	\subfloat{\includegraphics[width=0.24\linewidth]{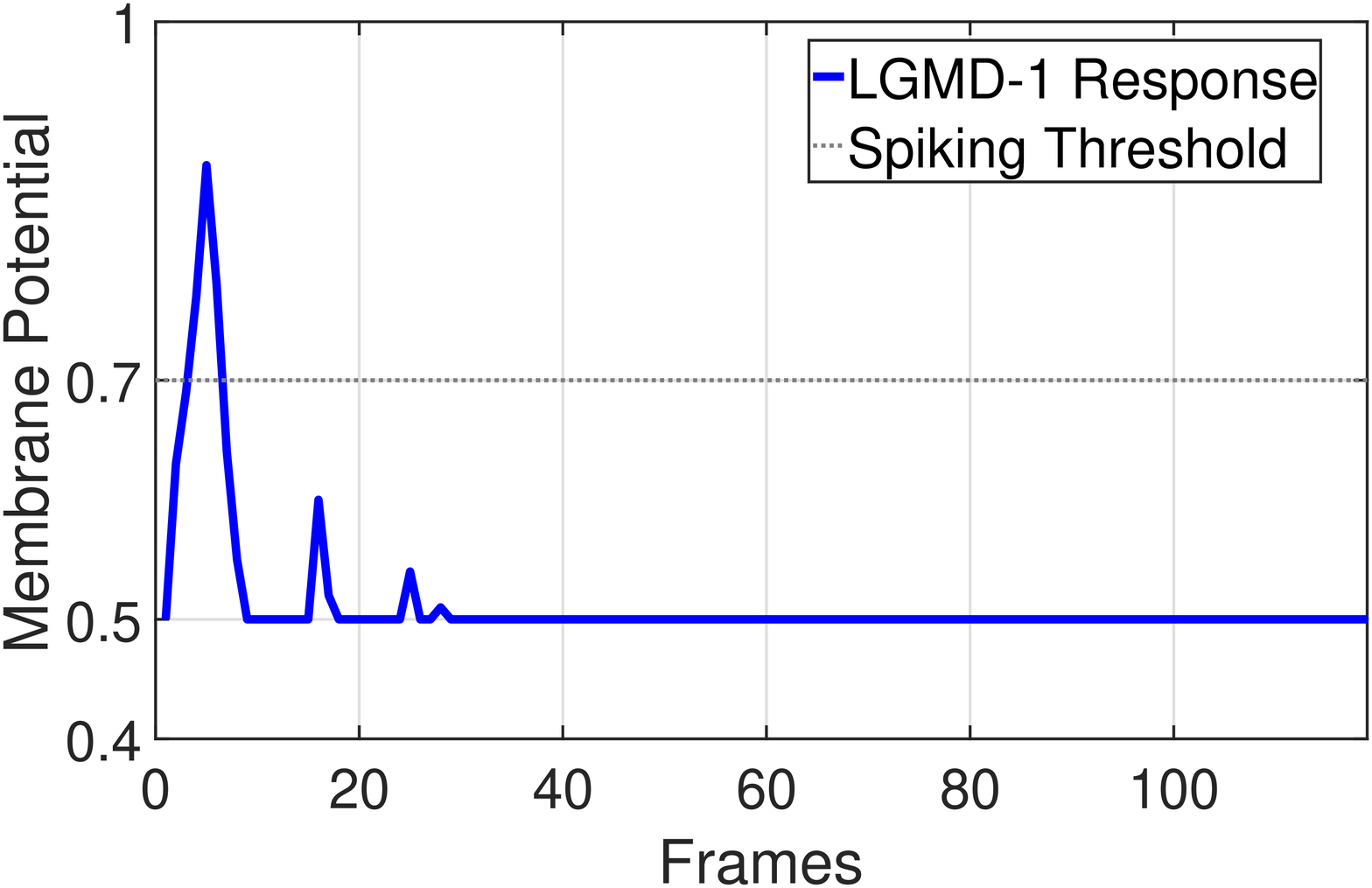}
		\label{Fig: receding-lgmd1}}
	\hfil
	\subfloat{\includegraphics[width=0.24\linewidth]{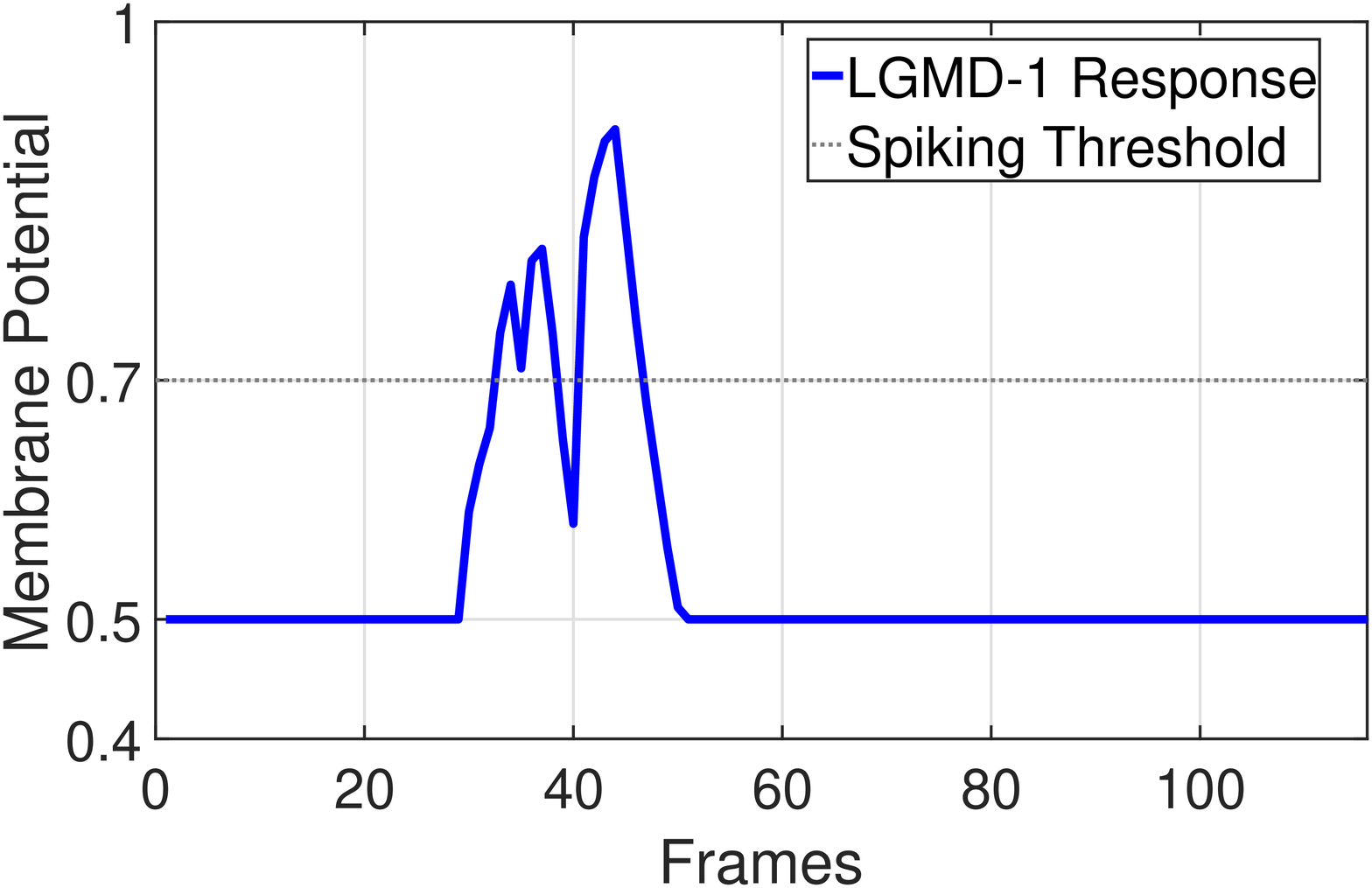}
		\label{Fig: transR-lgmd1}}
	\hfil
	\subfloat{\includegraphics[width=0.24\linewidth]{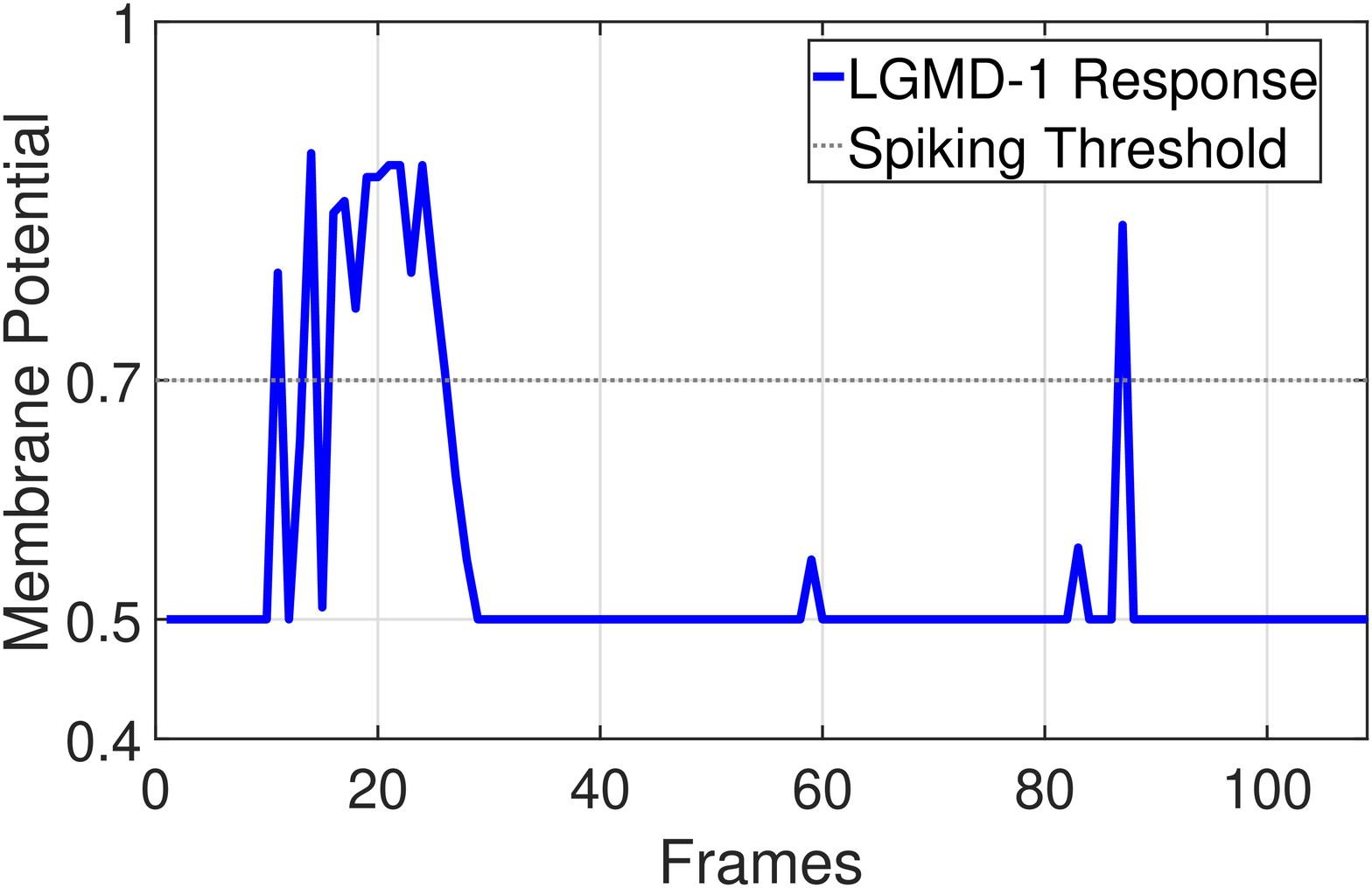}
		\label{Fig: transL-lgmd1}}
	\vfill
	\vspace{-0.1in}
	\subfloat{\includegraphics[width=0.24\linewidth]{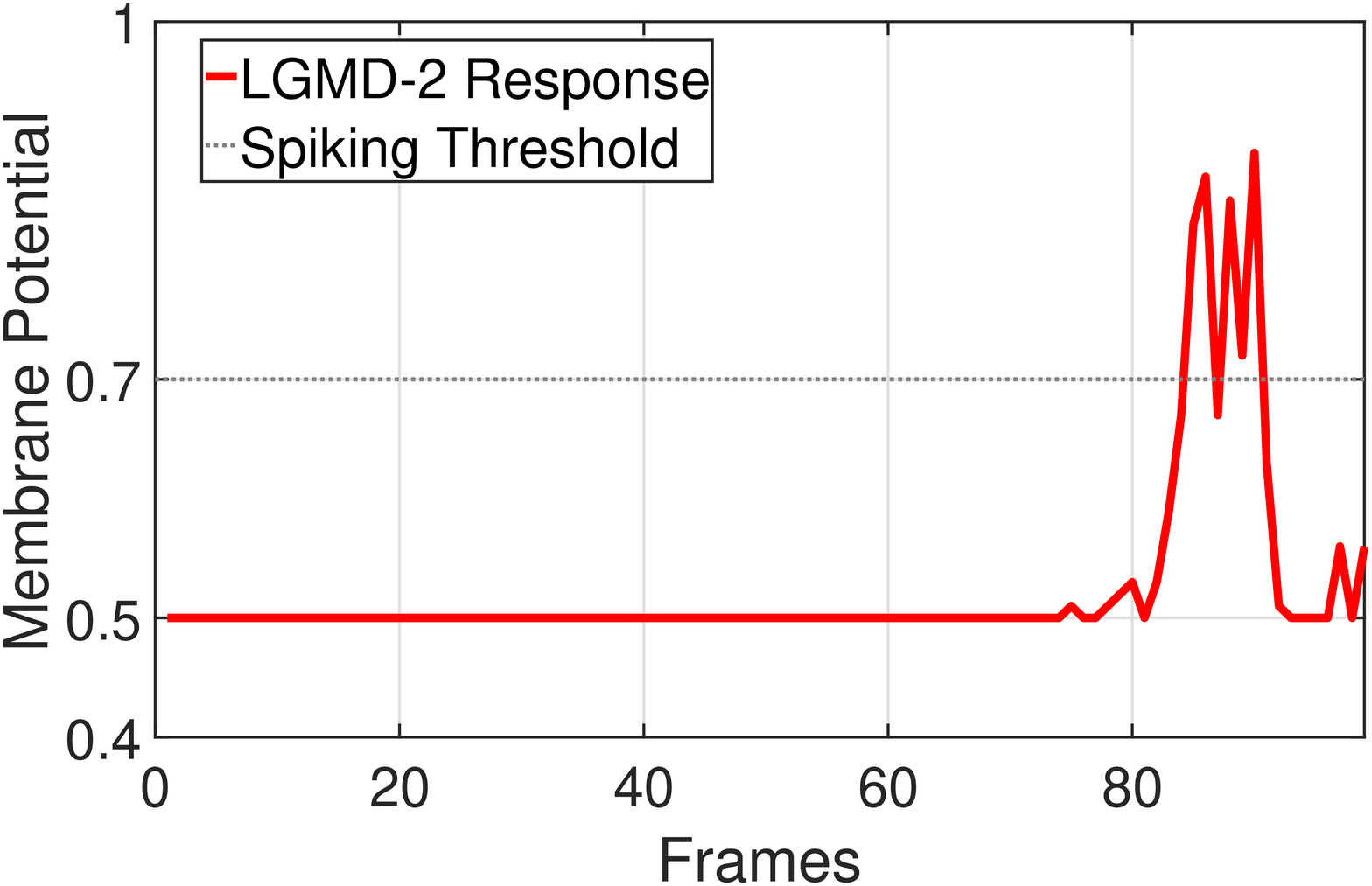}
		\label{Fig: approaching-lgmd2}}
	\hfil
	\subfloat{\includegraphics[width=0.24\linewidth]{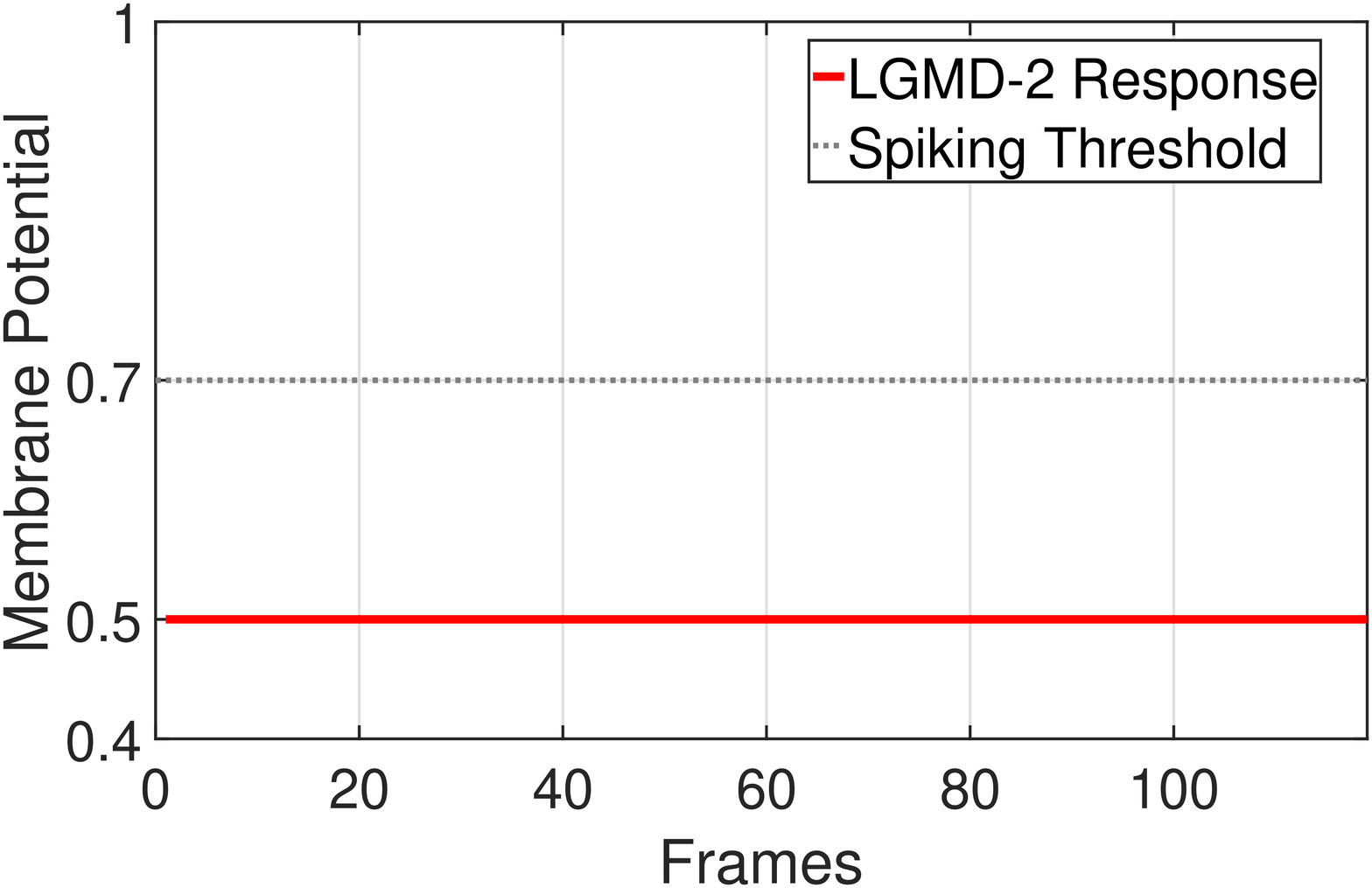}
		\label{Fig: receding-lgmd2}}
	\hfil
	\subfloat{\includegraphics[width=0.24\linewidth]{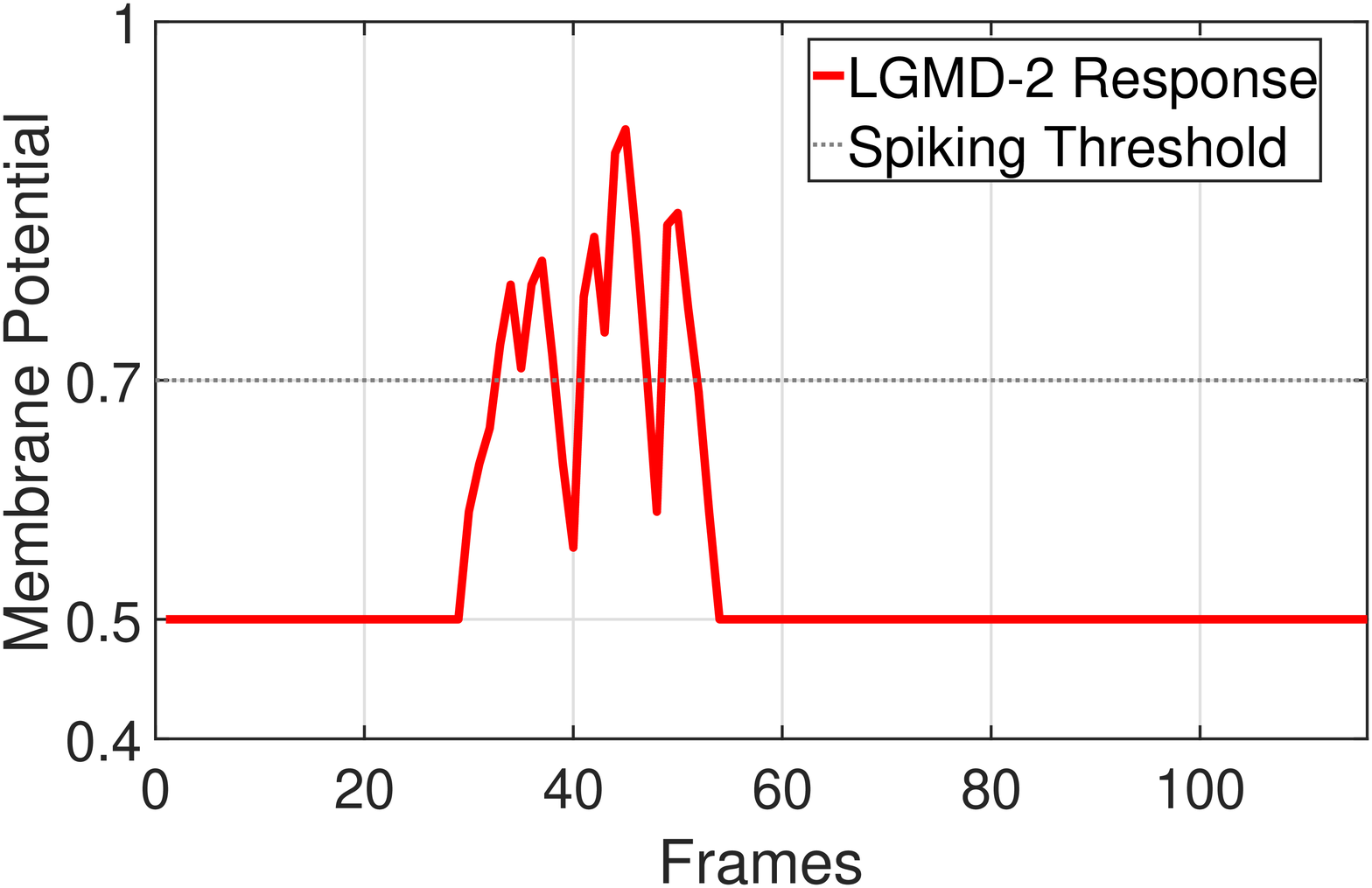}
		\label{Fig: transR-lgmd2}}
	\hfil
	\subfloat{\includegraphics[width=0.24\linewidth]{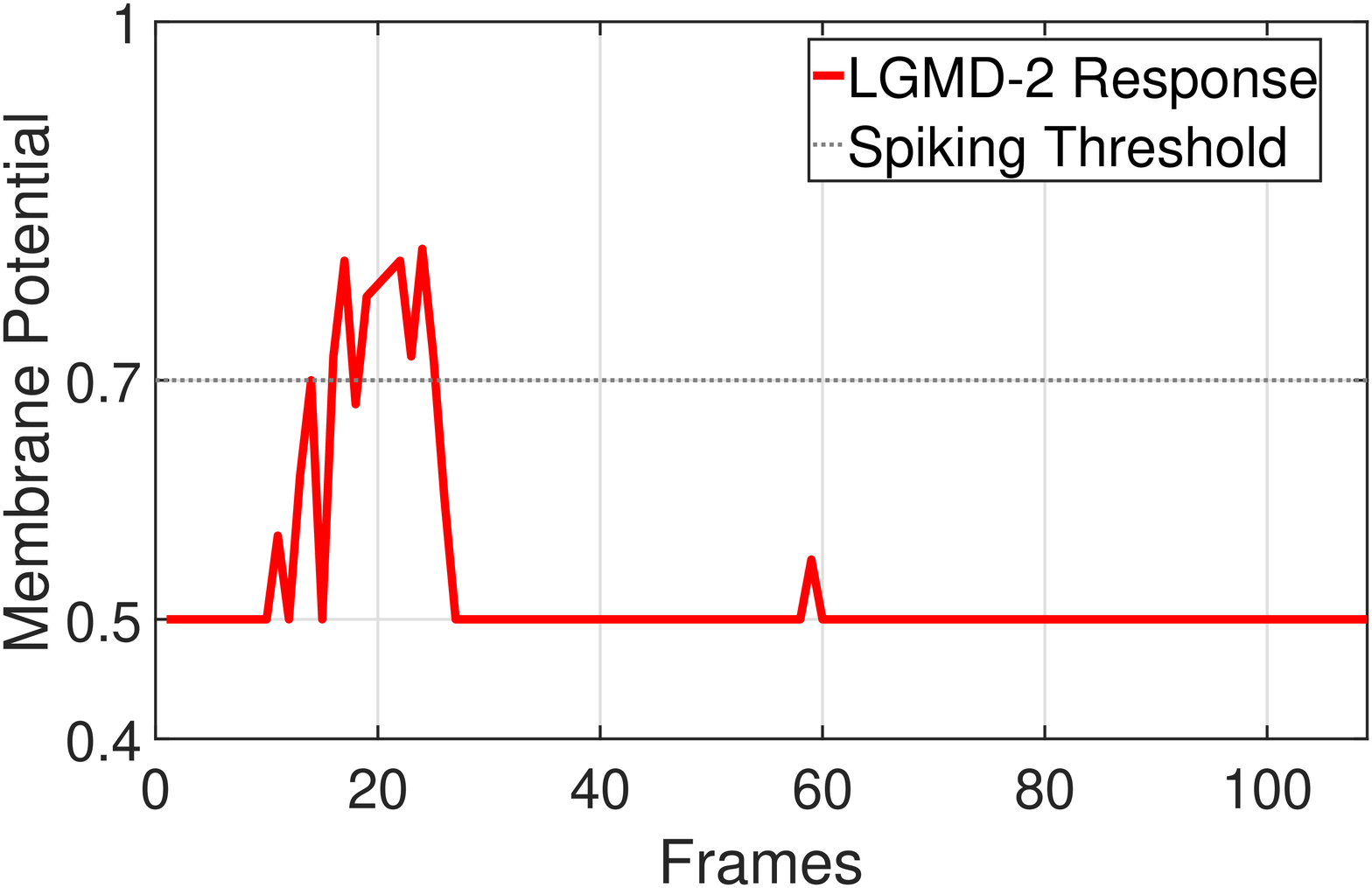}
		\label{Fig: transL-lgmd2}}
	\vfill
	\vspace{-0.1in}
	\subfloat{\includegraphics[width=0.24\linewidth]{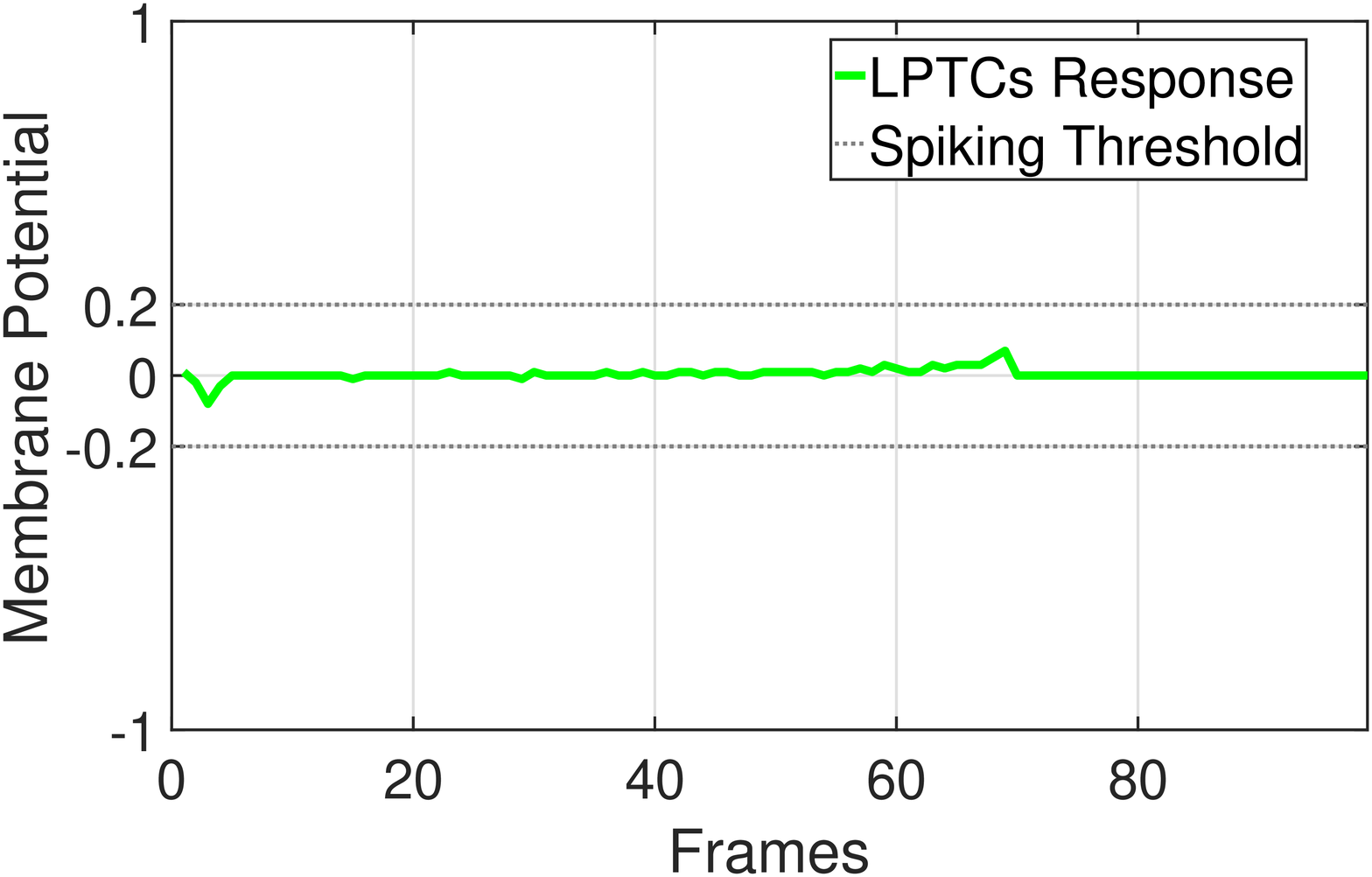}
		\label{Fig: approaching-dsn}}
	\hfil
	\subfloat{\includegraphics[width=0.24\linewidth]{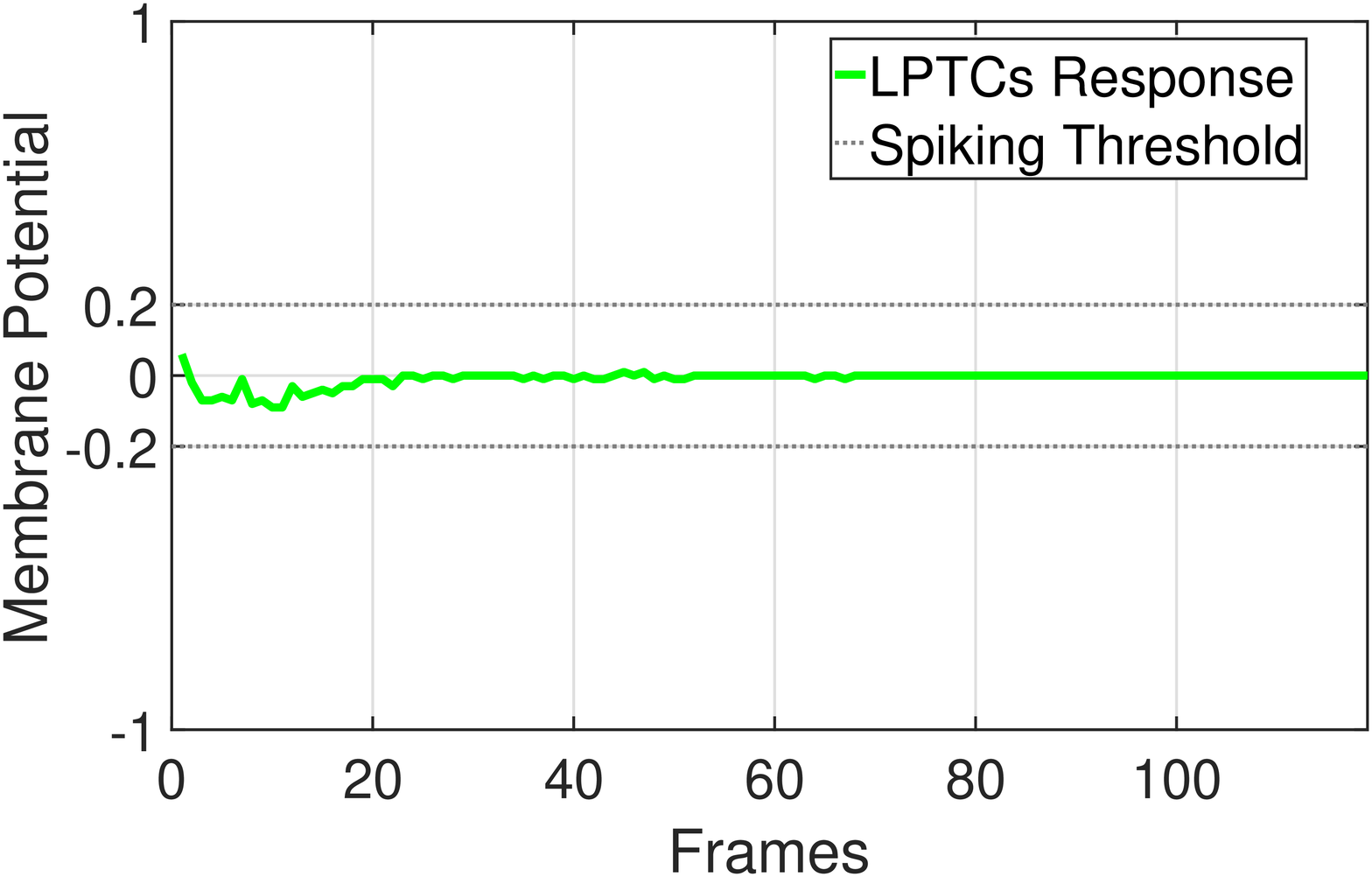}
		\label{Fig: receding-dsn}}
	\hfil
	\subfloat{\includegraphics[width=0.24\linewidth]{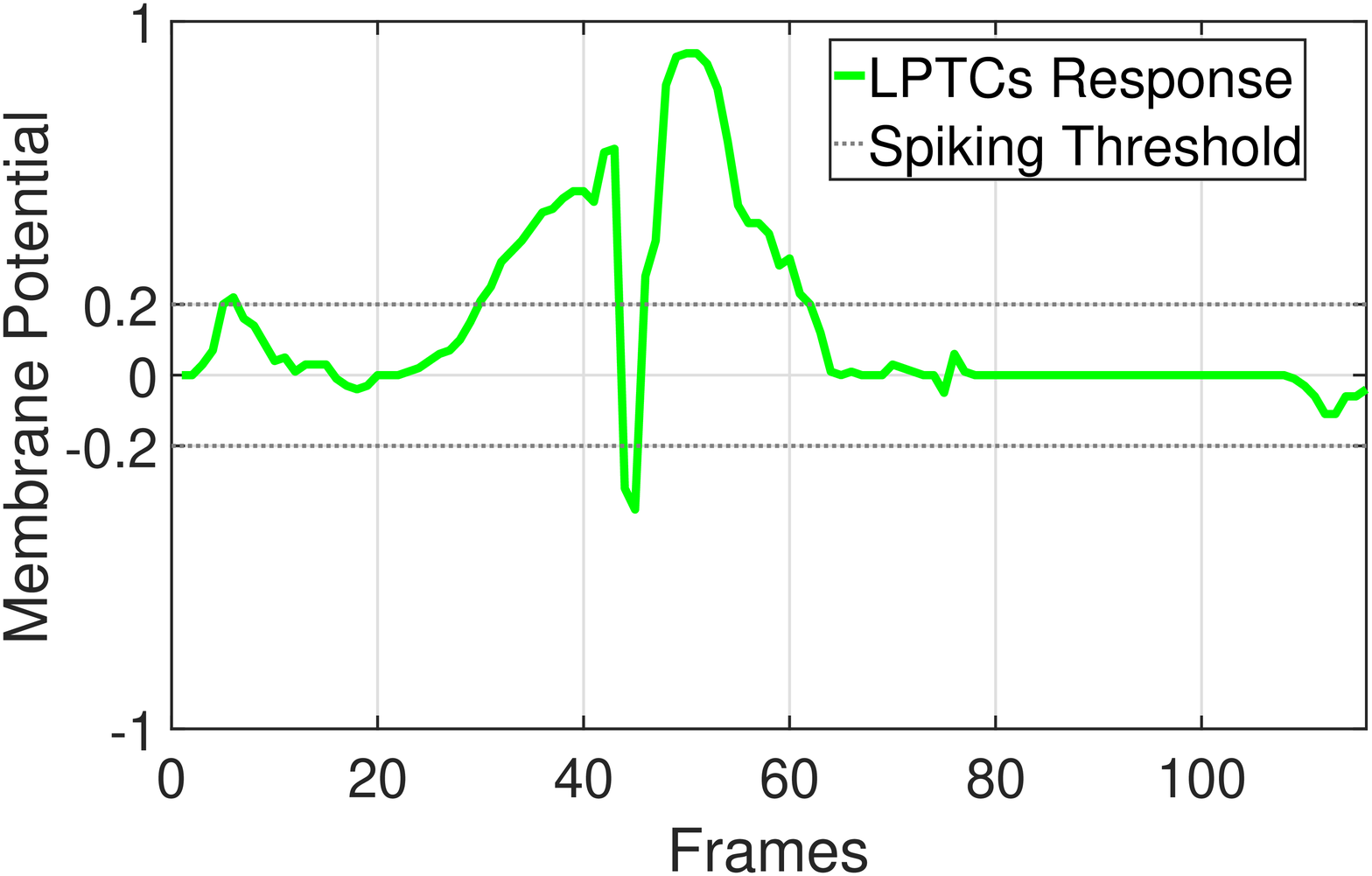}
		\label{Fig: transR-dsn}}
	\hfil
	\subfloat{\includegraphics[width=0.24\linewidth]{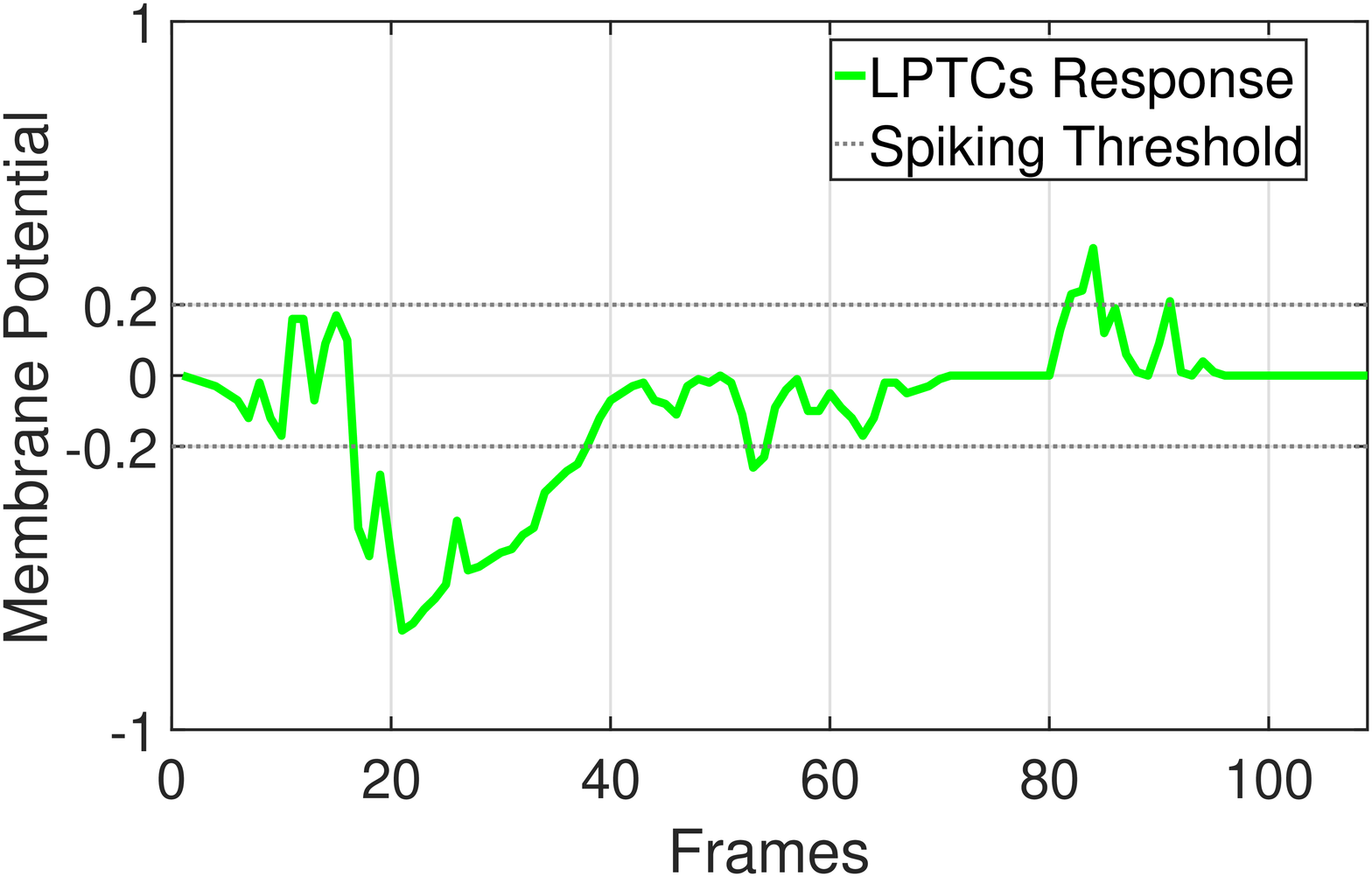}
		\label{Fig: transL-dsn}}
	\caption{
		Neuronal membrane potential challenged by four typical motion patterns. 
		Horizontal dashed lines indicate the corresponding spiking thresholds. 
		The stimulated robot is still.
	}
	\label{Fig: neuron-membrane-potential}
\end{figure*}

\begin{figure*}[t!]
	\vspace{-10pt}
	\centering
	\subfloat{\includegraphics[width=0.24\linewidth]{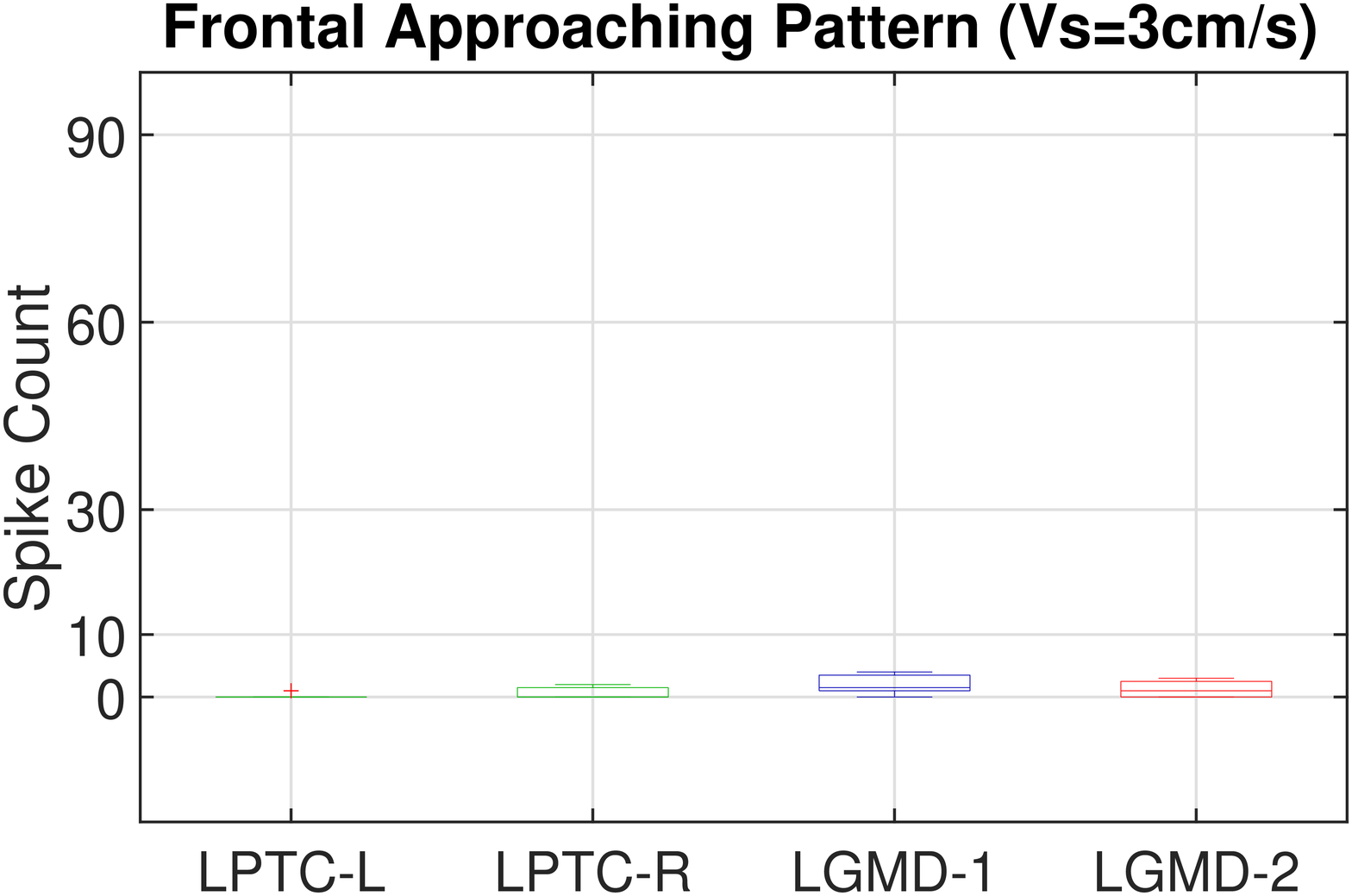}
		\label{Fig: approaching-s40-box}}
	\hfil
	\subfloat{\includegraphics[width=0.24\linewidth]{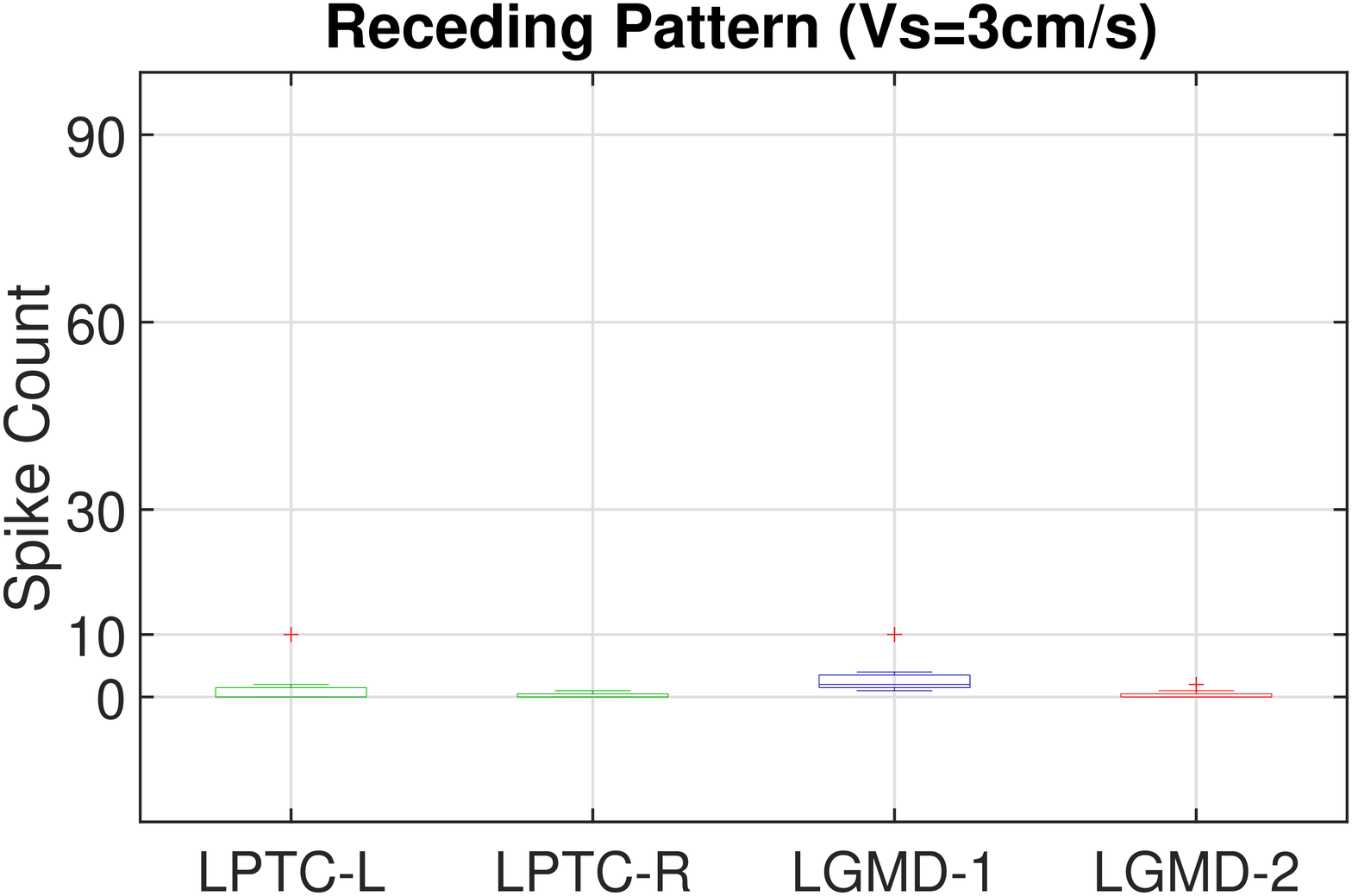}
		\label{Fig: receding-s40-box}}
	\hfil
	\subfloat{\includegraphics[width=0.24\linewidth]{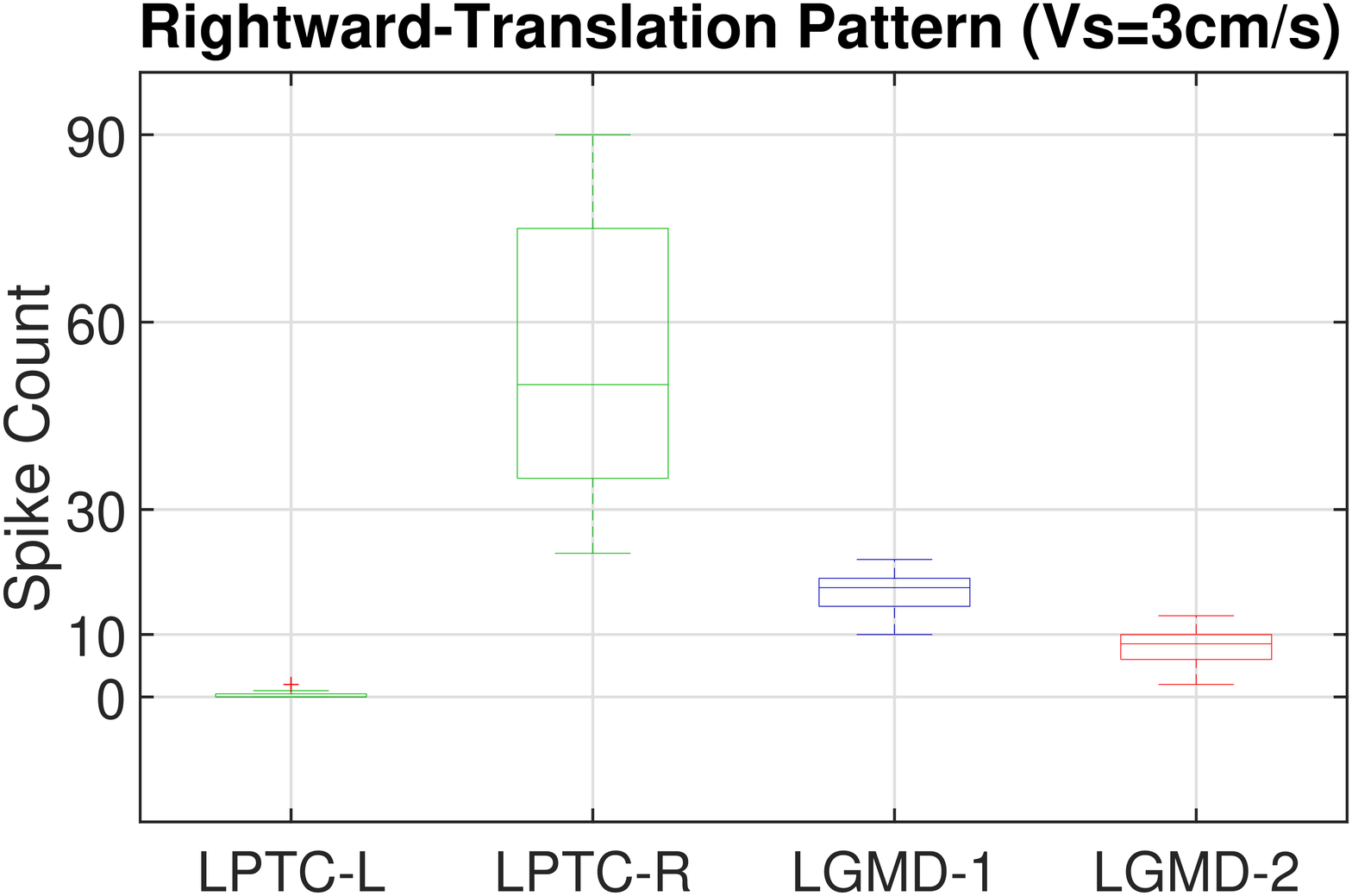}
		\label{Fig: translating-rightward-s40-box}}
	\hfil
	\subfloat{\includegraphics[width=0.24\linewidth]{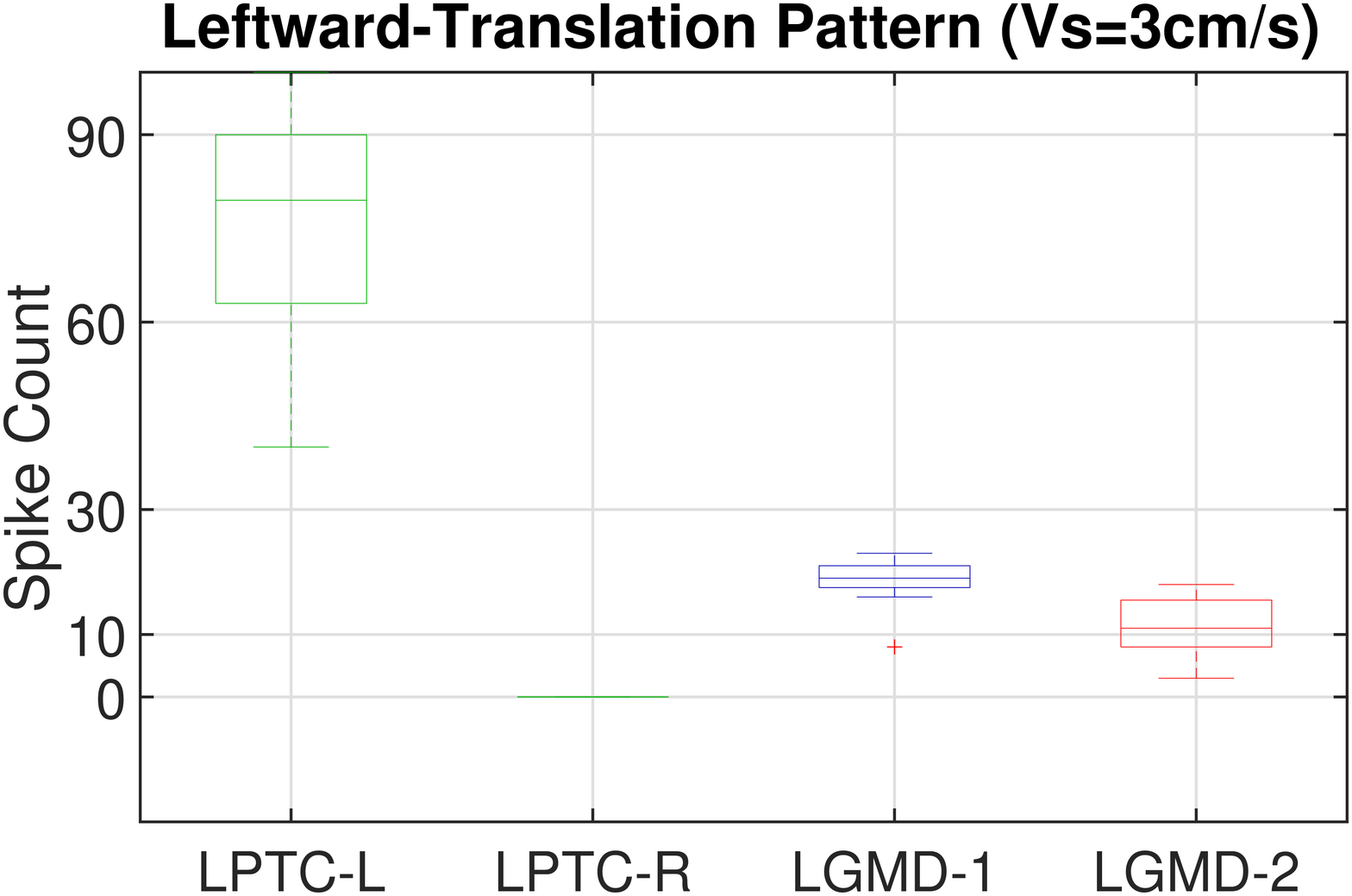}
		\label{Fig: translating-leftward-s40-box}}
	\vfill
	\vspace{-0.1in}
	\subfloat{\includegraphics[width=0.24\linewidth]{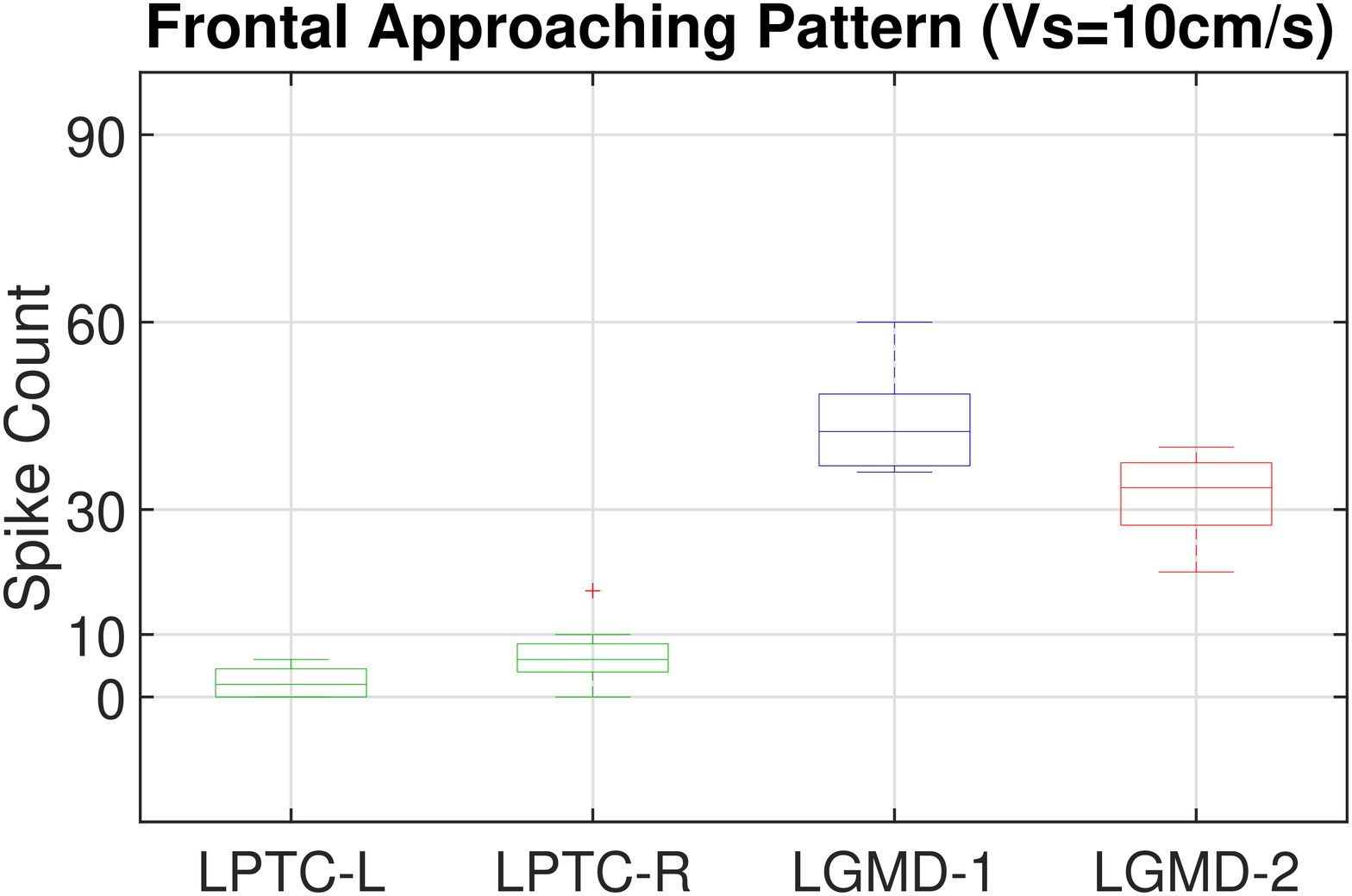}
		\label{Fig: approaching-s80-box}}
	\hfil
	\subfloat{\includegraphics[width=0.24\linewidth]{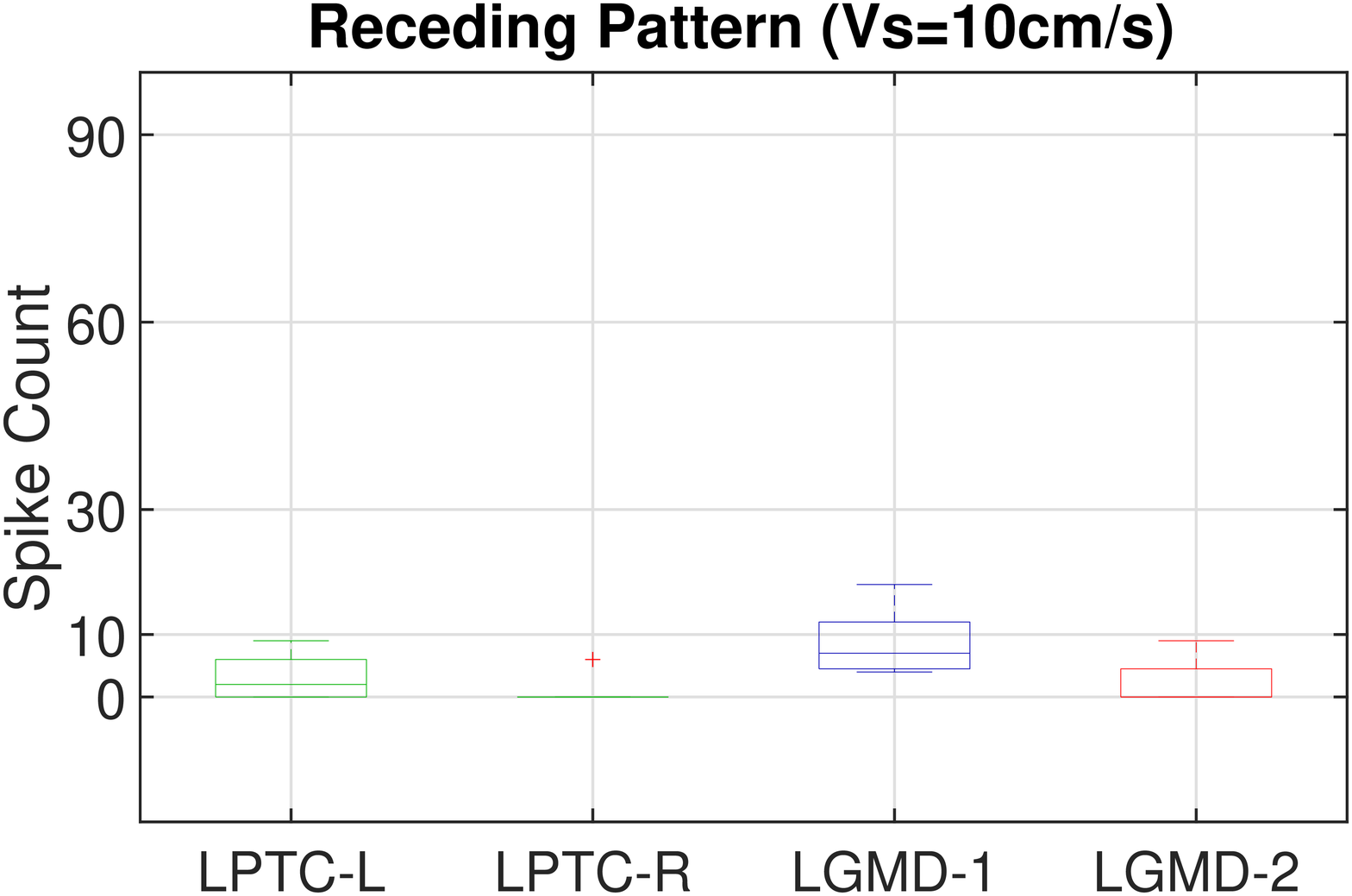}
		\label{Fig: receding-s80-box}}
	\hfil
	\subfloat{\includegraphics[width=0.24\linewidth]{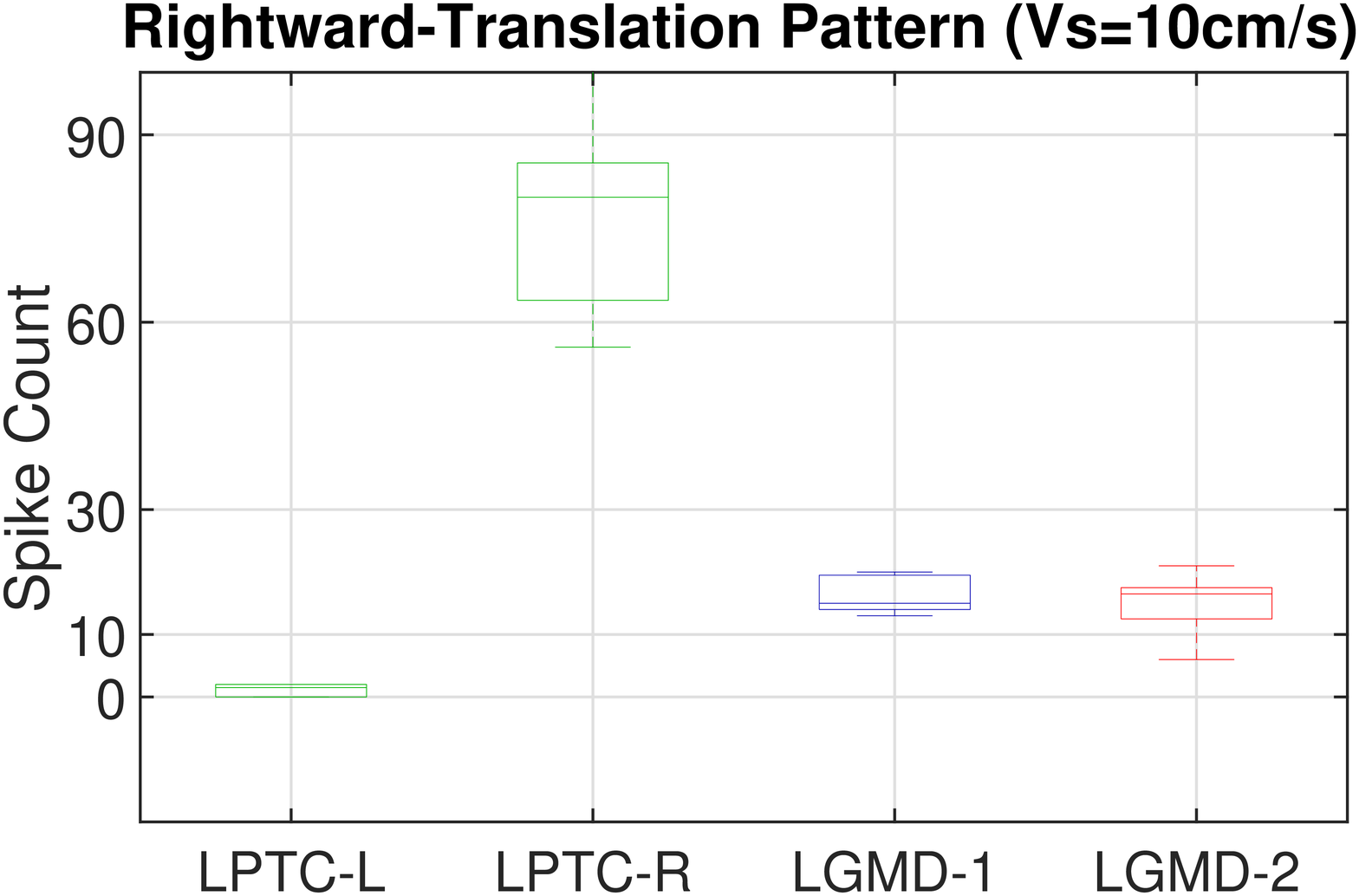}
		\label{Fig: translating-rightward-s80-box}}
	\hfil
	\subfloat{\includegraphics[width=0.24\linewidth]{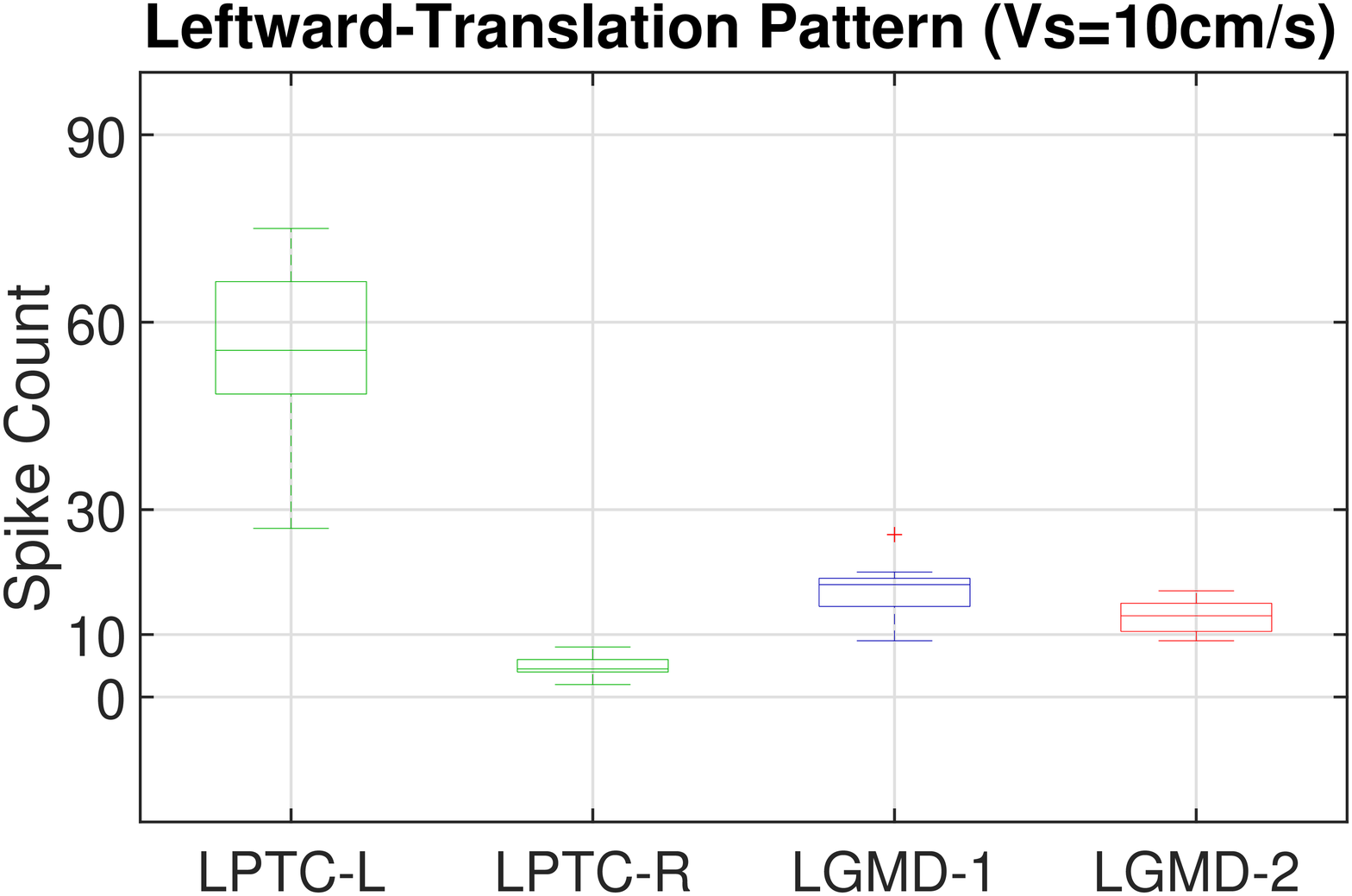}
		\label{Fig: translating-leftward-s80-box}}
	\vfill
	\vspace{-0.1in}
	\subfloat{\includegraphics[width=0.24\linewidth]{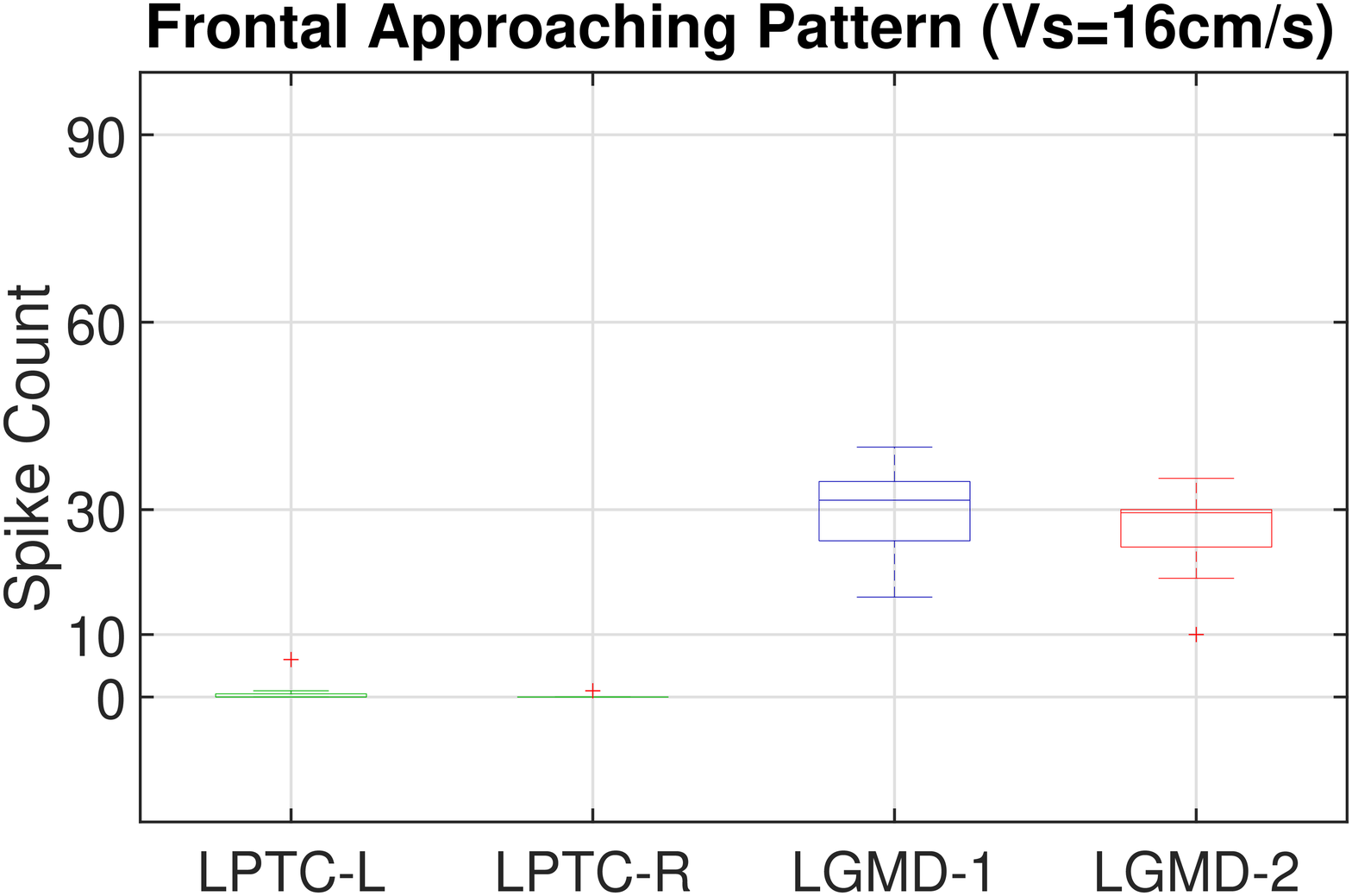}
		\label{Fig: approaching-s120-box}}
	\hfil
	\subfloat{\includegraphics[width=0.24\linewidth]{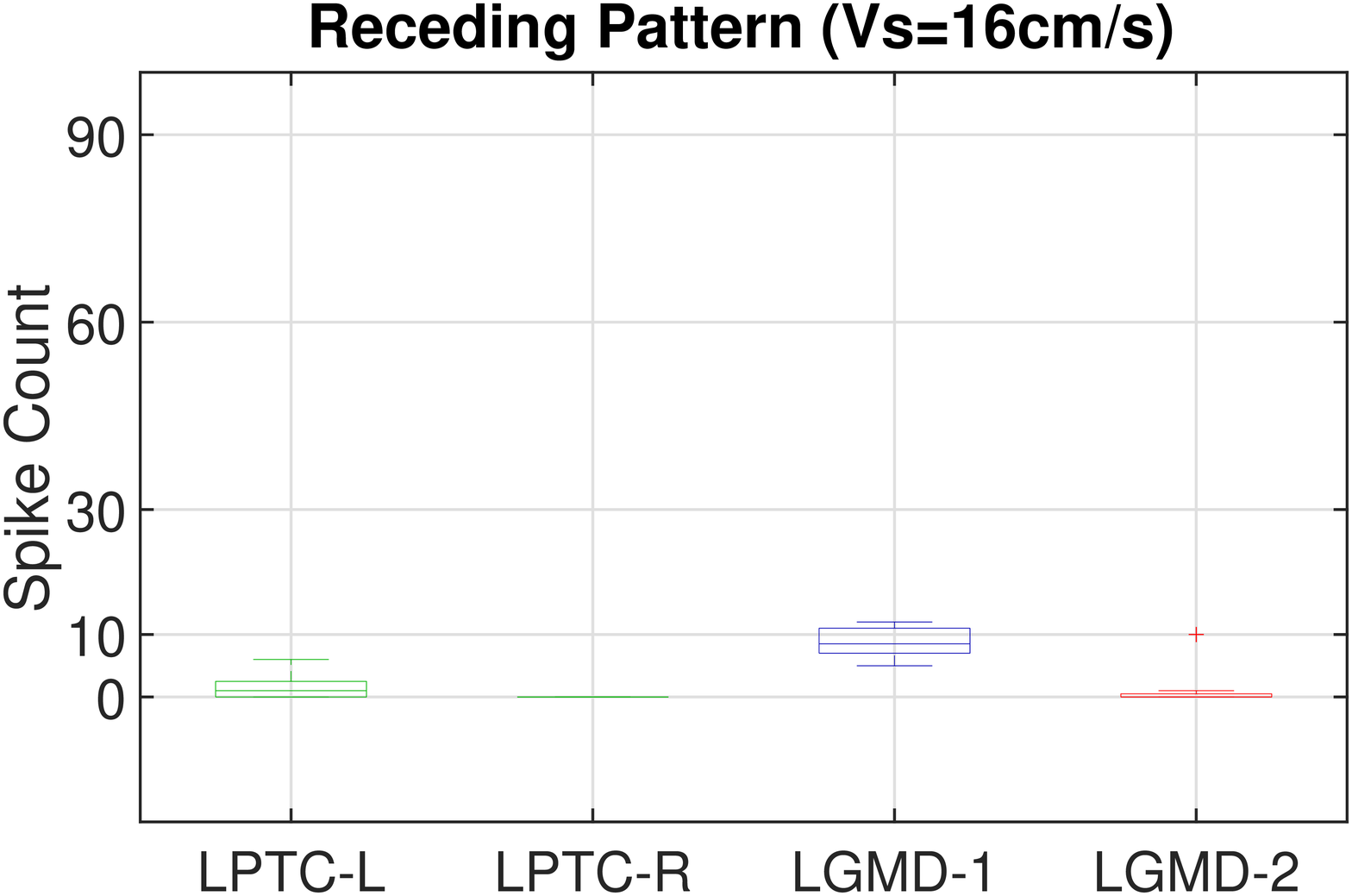}
		\label{Fig: receding-s120-box}}
	\hfil
	\subfloat{\includegraphics[width=0.24\linewidth]{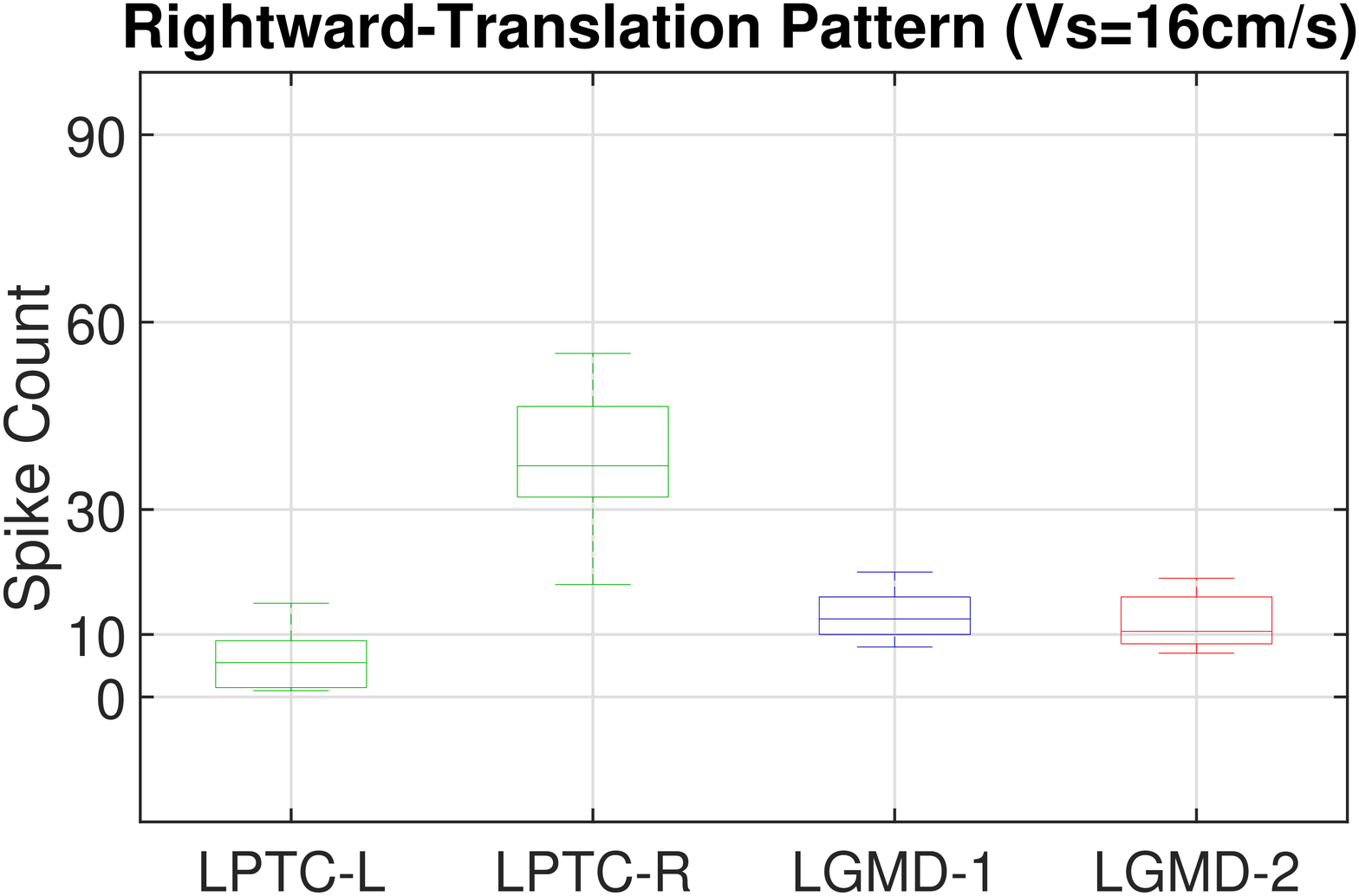}
		\label{Fig: translating-rightward-120-box}}
	\hfil
	\subfloat{\includegraphics[width=0.24\linewidth]{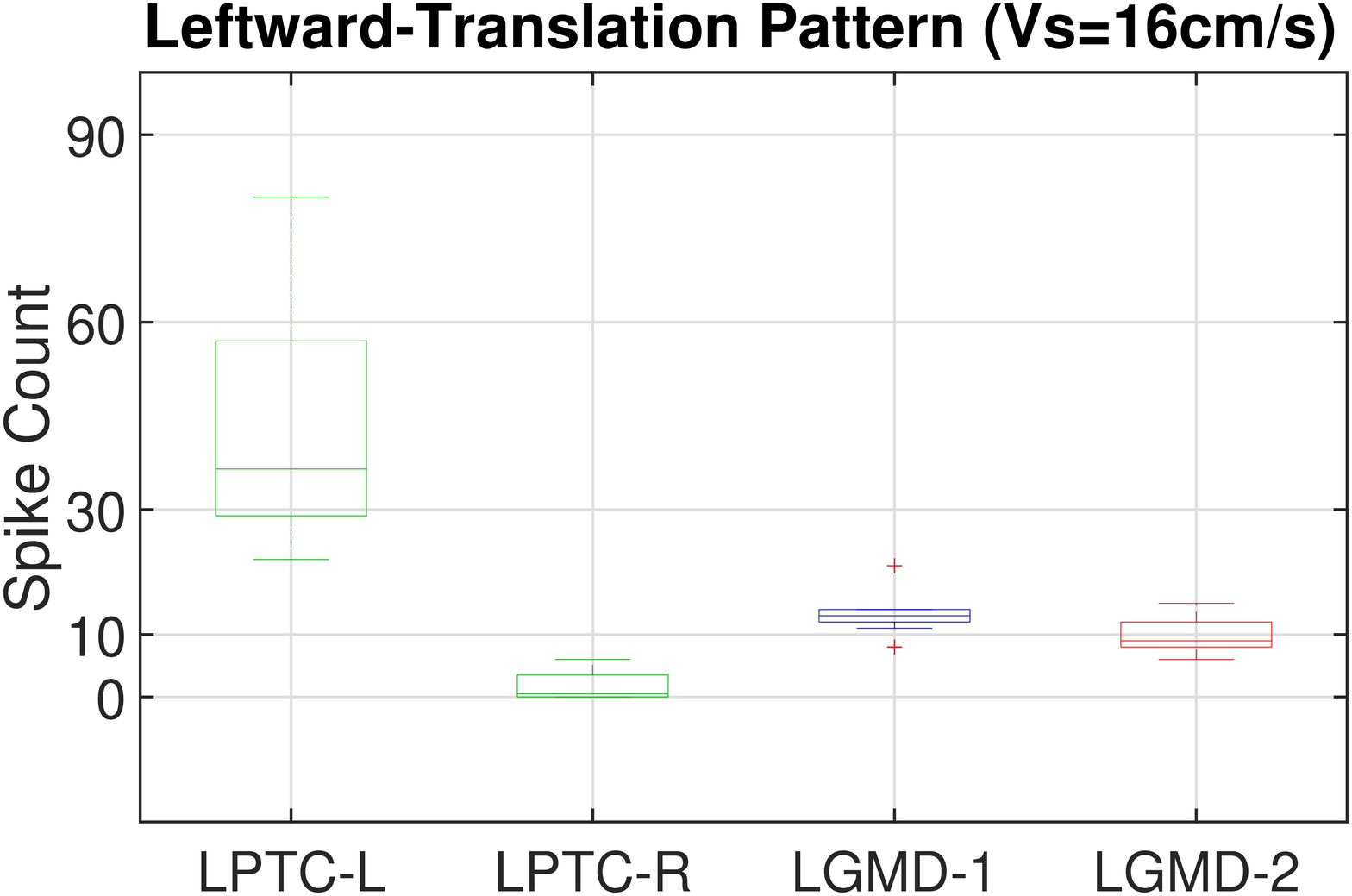}
		\label{Fig: translating-leftward-s120-box}}
	\caption{
		Statistical results of neuronal activation rate by spikes during every course of four basic motion patterns in Fig. \ref{Fig: neuron-membrane-potential} against three stimulus speeds, respectively, each of which was repeated ten times of tests. 
		The spikes during each course with an approximately identical time window are accumulated.
	}
	\label{Fig: neuron-activation-statistics}
	\vspace{-10pt}
\end{figure*}


\section{EXPERIMENTS}
\label{Sec: experiments}

Within this section, we present the multi-robot experiments and analyse the results. 
There are mainly two categories of tests: the open-loop and the closed-loop multi-robot arena tests. 
In the former kind of tests, we will firstly illustrate the outputting membrane potential of four neurons against typical motion patterns, in order to demonstrate the specific complementary DS between different modelled neurons. 
We will then systematically investigate the activations of distinct neurons by spike rates challenged by stimulus at different certain speeds, from slow to fast. 
Finally in the open-loop tests, we will demonstrate the neuronal activations against angular and frontal approaching objects. 
The moving stimulus was a \textit{Colias} robot in the arena.

In the second type of tests, multiple robots, each possessing the proposed model as the only collision-detecting sensor, interacted within the small arena at two densities (4 and 7 agents), respectively. 
To stress the improvement against translating motion interference in the collision sensing-and-avoidance task, we also compared its competence with two previous methods \cite{Fu-2016(LGMD2-BMVC),Fu2017a(LGMDs-IROS)}, under the same experimental settings. 
The statistical results are given in Table \ref{Tab: sr-table}.

\subsection{Open-loop Tests}

Firstly in the open-loop tests, Fig. \ref{Fig: neuron-membrane-potential} exemplifies the membrane potential of four modelled neurons stimulated by four typical motion patterns in dynamic visual scenes: frontal approaching, receding, nearby translating. 
Intuitively, the LGMDs neurons respond most strongly to the frontal approaching stimuli. 
The LGMD-1 responds briefly to the receding movements; whilst the LGMD-2 is rigorously suppressed by the recession. 
Conversely, the LPTCs are quiet to movements in depth, but highly activated by the translating stimuli, in which the LPTC-R responds to the rightward translation with positive response, while the LPTC-L has responsive preference to the leftward translation with negative response. 
The non-linear mapping in the pre-synaptic medulla processing of LPTCs (Eq. \ref{Eq: DSNs-emd}) achieves the specific DS. 
Note that the LGMDs are also activated by the nearby translations in either directions, which reflect the existing challenge to the current LGMDs models: the strong translating motion is likely recognised as potential collisions. 
The model responses have articulated the complementary functionality and selectivity between the four modelled neurons.

Next, Fig. \ref{Fig: neuron-activation-statistics} illustrates the neuronal activation by spikes during the four basic kinds of movements through repeated tests. 
The LPTCs represent much higher spike rate compared to the LGMDs when challenged by the translations at all tested speeds; whilst they are rigorously inhibited by the approaching and the receding stimuli. 
On the other hand, the LGMDs spike at high frequency by both the faster frontal approaching and nearby translations, responding most strongly to the approaching stimuli. 
Only the LGMD-1 neuron spikes by the recession. 
The four modelled neurons perform consistently at all tested speeds of the moving stimulus.

\begin{figure}[t!]
	\centering
	\subfloat[set-up]{\includegraphics[width=0.24\textwidth]{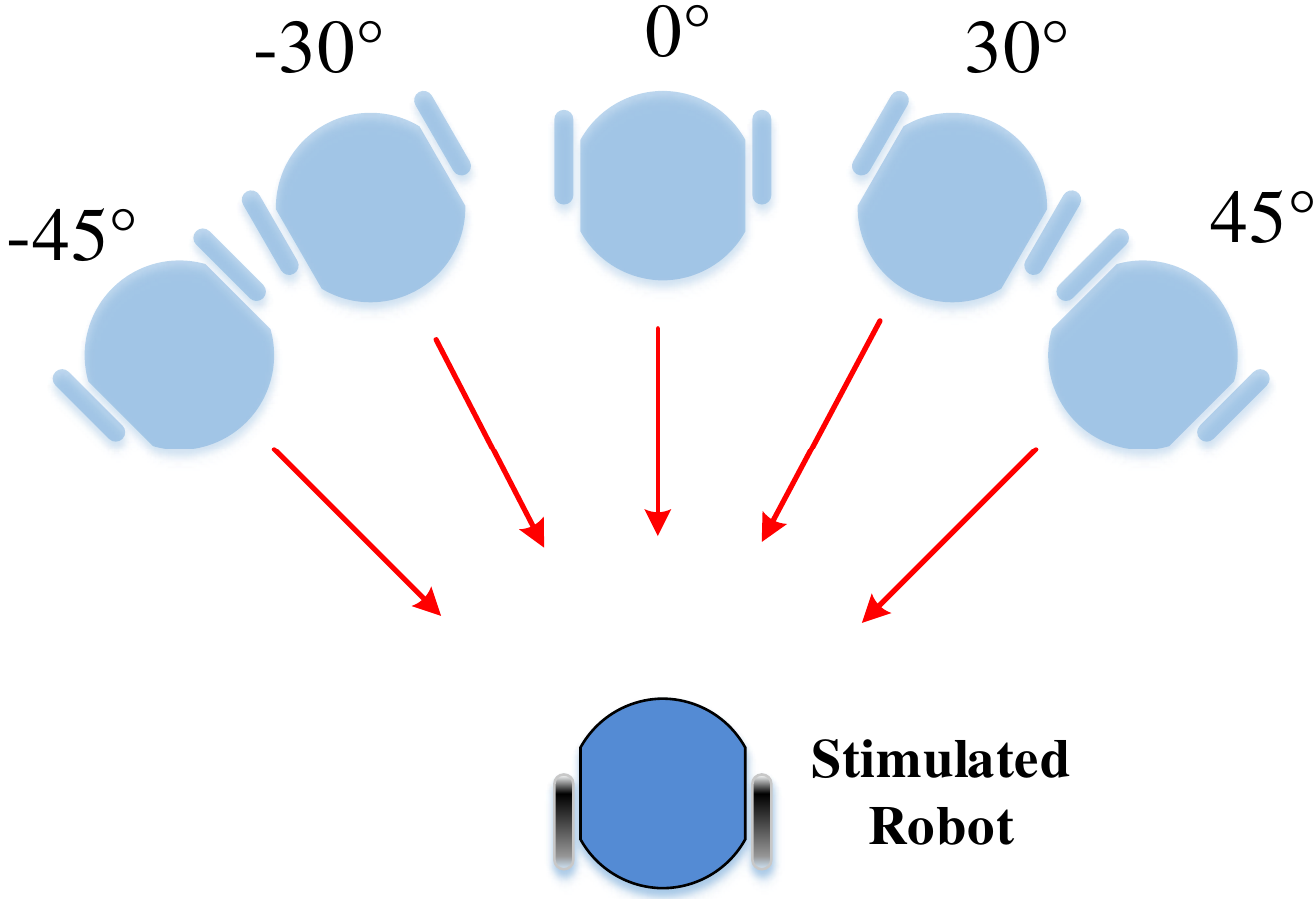}
		\label{Fig: aa-setup}}
	\hfil
	\subfloat[results]{\includegraphics[width=0.23\textwidth]{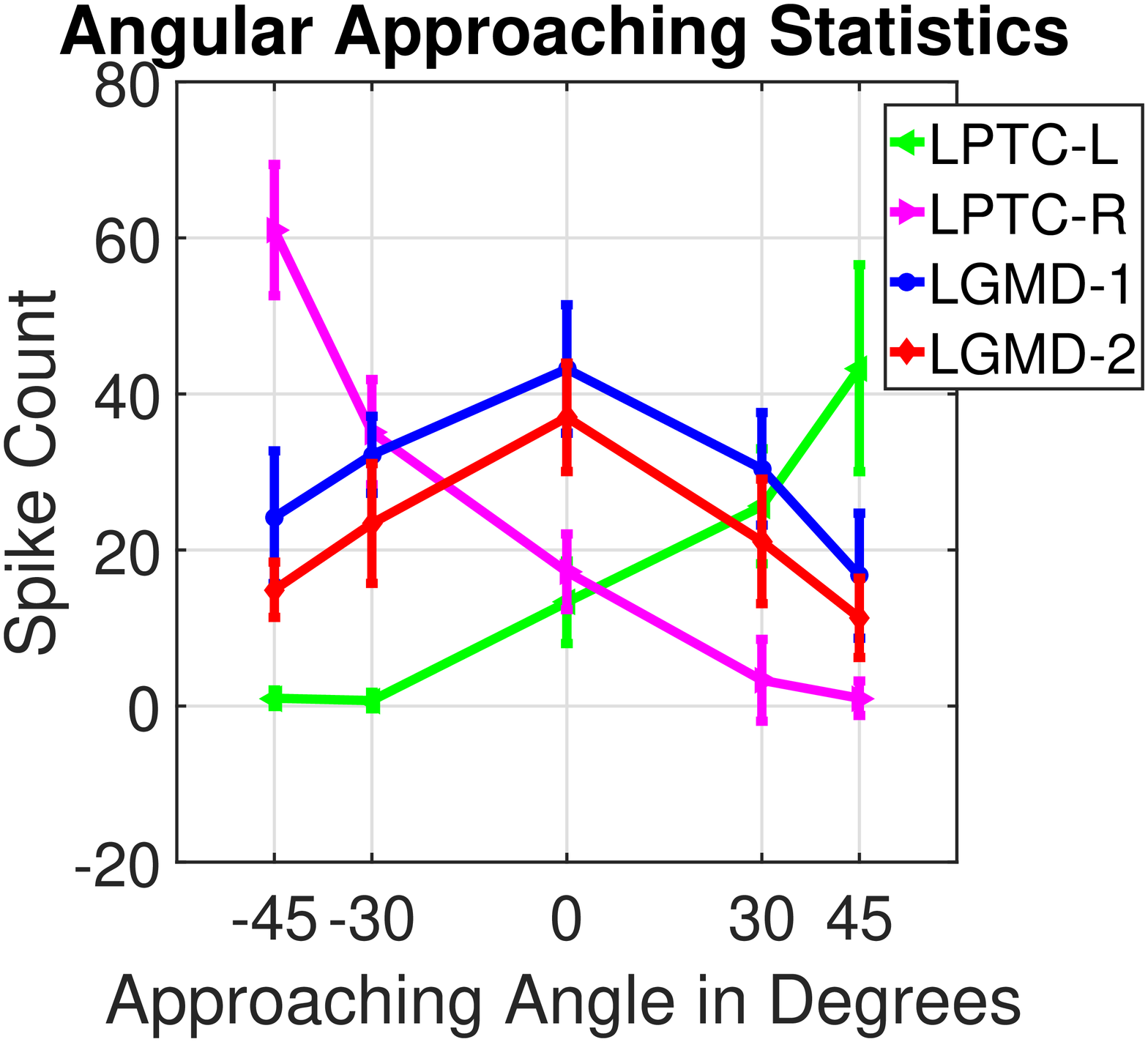}
		\label{Fig: aa-statistics}}
	\caption{
		Statistical results of robot angular approaching tests.
	}
	\label{Fig: angular-approach-tests}
	\vspace{-10pt}
\end{figure}

From our previous studies, we have observed that the angular approaching movements are very frequently challenging collision-detecting visual systems which should separate it well from the frontal approaching stimuli with different responsive preferences. 
In this research, to show the complementary functionality of four neuronal systems, we have also investigated the model responses against angular approaching stimuli. 
Fig. \ref{Fig: aa-setup} shows the testing set-up. 
Fig. \ref{Fig: aa-statistics} demonstrates the statistical results. 
Compared to the LGMDs, the LPTCs are more sensitive to the angular approaching from large angles. 
More precisely, the angular approaching stimulus from the left side brings about rightward translating (or elongating edges) that excites the LPTC-R; whilst the angular approaching from the right side activates the LPTC-L. 
By contrast, both the LGMDs spike most frequently to frontal over angular approaching stimuli showing higher spike frequency than the LPTCs. 
Accordingly, the complementary selectivity has been clearly shown.

The above experimental results have demonstrated the advantage of the proposed method for collision sensing. 
More specifically, the coordination between the LGMD-1 and the LGMD-2 works effectively to exclude the receding stimuli interference. 
The competition between the LPTCs and the LGMDs can further enhance the the proximity feature by frontal approaching objects against translating stimuli interference: the activation of LPTCs will rigorously inhibit the LGMDs. 
Therefore, the proposed method takes benefits from each specific motion sensitive neuron to focus on extracting merely frontal colliding features from dynamic environments.

\subsection{Comparative Arena Tests}
 
To verify the robustness of the proposed method in dynamic scenes, we have compared its competence with two previous single-type neuron computation methods: an LGMD-2 model in \cite{Fu-2016(LGMD2-BMVC)} and a hybrid LGMDs model in \cite{Fu2017a(LGMDs-IROS)}.

Multiple \textit{Colias} robots ran together in the arena at random linear speeds varying between 6 and 10cm/s, and at two densities (4 and 7) as introduced above, each lasting for one hour. 
More importantly, since more abundant motion features have been involved, we calculate the success rates (SRs) overtime based on a new criteria taking into account the translating interference, as presented in Table \ref{Tab: sr-table}. 

In case of collision avoidance to moving robot agents, the proposed approach shows much higher SR at both tested densities. 
The avoidance behaviours evoked by translating stimuli for other two models are prevented by the proposed model to a great extent, especially at the higher density. 
However, it also appears that the previous LGMD2 single-neuron model prevails in collision sensing when approaching the arena's peripheries. 
The reason is that the angular approaching to the walls could be suppressed by the activation of LPTCs in the proposed method.

\begin{table}[h!]
	\caption{Collision Sensing SRs in Arena Tests}
	\centering
	\begin{tabular}{|l|l|l|}
		\toprule
		\multicolumn{3}{c}{\textbf{Events}: Colliding with Robots, Peripheries (CR, CP)}\\
		\multicolumn{3}{c}{Avoiding Approaching, Translating Robots (AA, AT)}\\
		\multicolumn{3}{c}{Avoiding Arena Peripheries (AP)}\\
		\multicolumn{3}{c}{SR1=AP/(AP+CP)$\times$100\%}\\ 
		\multicolumn{3}{c}{SR2=AA/(AA+AT+CR)$\times$100\%}\\
		\cmidrule{1-3}
		4-Robots Scene					&SR1	   			&SR2\\
		\cmidrule{1-3}
		LGMD-2 only						&\textbf{96.7\%}	&80.0\%\\
		LGMD-1 \& LGMD-2				&88.1\%	   			&73.9\%\\
		\textbf{The proposed model}		&90.3\%				&\textbf{87.3\%}\\
		\cmidrule{1-3}
		7-Robots Scene					&SR1	   			&SR2\\
		\cmidrule{1-3}
		LGMD-2 only						&\textbf{95.0\%}	&75.2\%\\
		LGMD-1 \& LGMD-2				&81.7\%	   			&67.8\%\\
		\textbf{The proposed model}		&83.4\%				&\textbf{90.6\%}\\
		\bottomrule
	\end{tabular}
	\label{Tab: sr-table}
\end{table}


\section{CONCLUSION}
\label{Sec: conclusion}

In this paper, we have presented a novel complementary visual neuronal systems model with enhanced frontal collision selectivity in dynamic environments mixed with diverse motion patterns. 
Four wide-field movement sensitive neuronal modules, the LGMD-1, the LGMD-2, the LPTC-R and the LPTC-L with complementary functionality and selectivity, form a hybrid neural system model. 
With coordination and competition between different activated neurons, the proximity feature by frontal approaching objects rather than any other categories of movements has been largely sharpened up, which is the main contribution of this study. 
The multi-robot on-line experiments including open-loop and closed-loop tests have demonstrated the effectiveness and robustness of the proposed method that outperforms previous single-type neuron computation models against translational motion interference in collision perception.

This research also indicates several future works. 
One major concern should be put forth in trialling the model in more complex and cluttered scenes, the natural scenarios. 
Another interesting effort could be made on introducing this visual system into more mobile platforms including ground vehicles and UAVs. 
Moreover, the current model processes visual information in a feed-forward manner; for better adapting to more unpredictable environments, the learning methods with feedback loop will be carried on.


\bibliographystyle{IEEEtran}

\bibliography{qinbingbib}

\addtolength{\textheight}{-12cm}  
\end{document}